\documentclass{article} % For LaTeX2e
\usepackage{simplexai}

\usepackage[T1]{fontenc}
\usepackage{microtype}
\usepackage{hyperref}
\usepackage{url}
\usepackage{graphicx}
\usepackage{booktabs}
\usepackage{subcaption}
\usepackage{multirow}
\usepackage[export]{adjustbox}
\usepackage[table]{xcolor}
\usepackage{caption}
\usepackage{float}
\usepackage{enumitem}
\usepackage{amsmath}
\usepackage{mathtools}
\usepackage{amsthm}
\usepackage{wrapfig}

\usepackage{tabularx}
\usepackage{ragged2e}
\usepackage[most]{tcolorbox}
\usepackage{listings}
\usepackage{fontawesome}

\usepackage{iftex}
\ifPDFTeX
  \usepackage{CJKutf8}
  \newcommand{\zh}[1]{\begin{CJK*}{UTF8}{gbsn}#1\end{CJK*}}
\else
  \usepackage[UTF8]{ctex}
  \newcommand{\zh}[1]{#1}
\fi

\newtcolorbox{promptbox}[1]{
    enhanced,
    breakable,
    colback=blue!3!white,
    colframe=black!70,
    coltext=black,
    fonttitle=\bfseries,
    coltitle=black,
    colbacktitle=gray!10!white,
    title=#1,
    boxrule=0.8pt,
    arc=4pt,
    fontupper=\small,
    left=2mm, right=2mm, top=2mm, bottom=2mm,
}
\definecolor{trajsearch}{HTML}{1B7F4B}
\definecolor{trajvisit}{HTML}{C2410C}
\definecolor{trajthink}{HTML}{1D4ED8}
\definecolor{trajanswer}{HTML}{6D28D9}
\newcommand{\trtag}[2]{\par\smallskip\noindent\makebox[4em][l]{\textcolor{#1}{\sffamily\scriptsize\bfseries #2}}}
\newcommand{\trsearch}[1]{\trtag{trajsearch}{SEARCH}#1}
\newcommand{\trvisit}[1]{\trtag{trajvisit}{VISIT}#1}
\newcommand{\trthink}[1]{\trtag{trajthink}{THINK}#1}
\newcommand{\transwer}[1]{\trtag{trajanswer}{ANSWER}#1}

\newcommand{\trquote}[1]{\par\smallskip{\leftskip=1.6em\rightskip=0.5em\small\color{black!72}\textcolor{trajthink}{\sffamily\scriptsize\bfseries thought:}\ \itshape #1\par}\vskip 1pt}
\newcommand{\trturn}[1]{\par\smallskip\noindent{\sffamily\scriptsize\bfseries\color{simplexmuted}TURN #1}\par\vskip-1pt}
\newcommand{\trobs}[1]{\trtag{simplexmuted}{$\hookleftarrow$\,obs}#1}
\newtcolorbox{trajbox}[1]{
    enhanced jigsaw, breakable,
    colback=gray!2!white, colframe=simplexink!70,
    fonttitle=\bfseries\sffamily\small, coltitle=white, colbacktitle=simplexink,
    title={#1}, boxrule=0.6pt, arc=3pt,
    left=2mm, right=2mm, top=1.5mm, bottom=1.5mm, fontupper=\small,
    title after break={\normalfont\itshape Trajectory (continued)},
    pad at break*=2mm,
}
\usepackage[capitalize,noabbrev]{cleveref}

\theoremstyle{plain}

\theoremstyle{definition}

\theoremstyle{remark}

\usepackage{lettrine}
\setlist{nosep} % Global compact lists

\usepackage[textsize=tiny]{todonotes}

\usepackage{tikz}
\usepackage{pgfplots}
\pgfplotsset{compat=1.18}

\definecolor{darkblue}{rgb}{0, 0, 0.5}
\hypersetup{colorlinks=true, citecolor=darkblue, linkcolor=darkblue, urlcolor=simplexaccentdeep}

\title{\textcolor{simplexaccentdeep}{LiteResearcher}: A Scalable Agentic RL Training Framework for Deep Research Agent}
\author{\mdseries
\begin{minipage}{\linewidth}
\raggedright
\textbf{Bince Qu}$^{1,2,*}$,\enspace
\textbf{Wanli Li}$^{1,2,*}$,\enspace
\textbf{Bo Pan}$^{1}$,\enspace
\textbf{Jianyu Zhang}$^{1}$,\enspace
\textbf{Zheng Liu}$^{3,\dagger}$,\enspace
\textbf{Pan Zhang}$^{2}$\\[2pt]
\textbf{Wei Chen}$^{1}$,\enspace
\textbf{Bo Zhang}$^{1,2,\dagger}$ \\[6pt]
$^1$\,Zhejiang University \quad $^2$\,Simplex AI \quad $^3$\,The Hong Kong Polytechnic University \\[3pt]
{\small \texttt{\{wanli\_li@zju.edu.cn, tonyzhang@simplexai.com\}}} \\[4pt]
{\small \faGithub~\href{https://github.com/simplexai-labs/LiteResearcher}{\texttt{https://github.com/simplexai-labs/LiteResearcher}}} \\[2pt]
{\small \raisebox{-0.1em}{\includegraphics[height=1em]{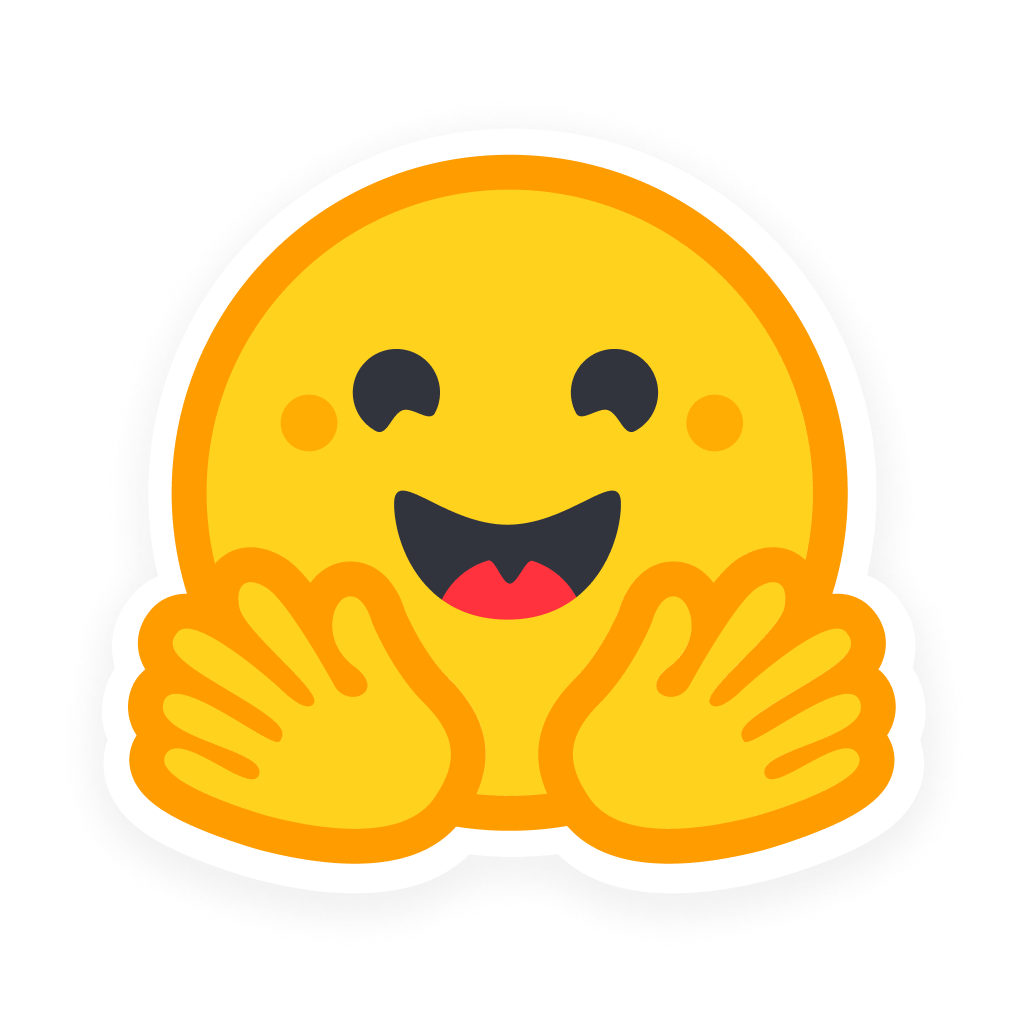}}~\href{https://huggingface.co/simplex-ai-inc/LiteResearcher-4B}{\texttt{https://huggingface.co/simplex-ai-inc/LiteResearcher-4B}}}
\end{minipage}
}
\begin{document}
\renewcommand{\thefootnote}{\fnsymbol{footnote}}
\footnotetext[1]{Equal contributors, in alphabetical order by first name.}
\footnotetext[2]{Corresponding authors.}

\runningtitle{LiteResearcher}
\maketitle
\thispagestyle{simplextitle}

\begin{abstract}
Reinforcement Learning (RL) has emerged as a powerful training paradigm for LLM-based agents. However, scaling agentic RL for deep research remains constrained by two coupled challenges: hand-crafted synthetic data fails to elicit genuine real-world search capabilities, and real-world search dependency during RL training introduces instability and prohibitive cost, which limits the scalability of Agentic RL. \textbf{LiteResearcher} is a training framework that makes Agentic RL scalable: by constructing a lite virtual world that mirrors real-world search dynamics, we enable a continuously improving training recipe that empowers a tiny search agent to outperform large-scale open-source and commercial models (e.g., Tongyi DeepResearch and Claude-4.5 Sonnet). Specifically, on common benchmarks such as GAIA and Xbench, our \textbf{LiteResearcher-4B} achieves open-source state-of-the-art results of \textbf{71.3\%} and \textbf{78.0\%} respectively, demonstrating that scalable RL training is a key enabler for Deep Research Agents. 
\end{abstract}

\begin{figure}[H]
    \vspace*{-1em}
    \makebox[\textwidth][c]{\includegraphics[width=1\textwidth]{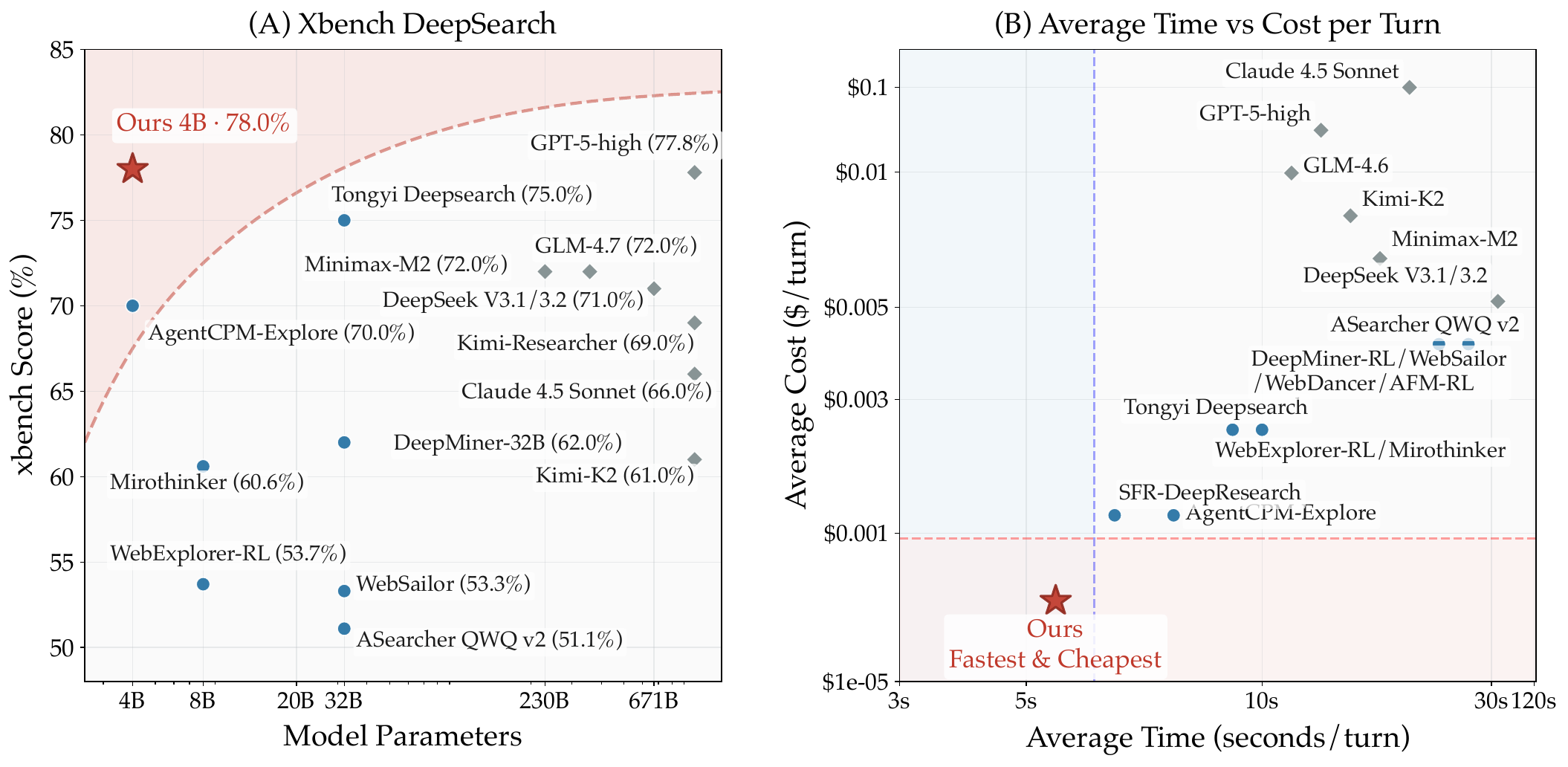}}
    \caption{\textbf{Performance of LiteResearcher.} \textit{Left:} Accuracy comparison on the Xbench DeepSearch benchmark across models of various scales. \textit{Right:} Average rollout latency and cost per turn.}
    \label{fig:performance_speed}
\end{figure}

\section{Introduction}
Recent advances in reinforcement learning (RL) on reasoning models~\citep{liu2025prorl,guo2025deepseek,team2025kimi} have demonstrated that continuously scaling model capabilities is achievable within stable training environments and with data of appropriate difficulty. However, when transitioning to agentic RL, where reasoning must be integrated with tool use, prior efforts have struggled to achieve similar sustained improvements. We identify the primary bottleneck as the inherent noise and instability of the real-world environment. In the domain of Coding Agents~\citep{wei2025swe,jain2025r2e}, this challenge has been effectively addressed through the use of isolated sandboxes configured to simulate production environments while filtering out infrastructure-level noise, thereby enabling scalable training.

However, existing works in the domain of Deep Research have yet to realize such scalable RL training. Online RL frameworks~\citep{zheng2025deepresearcher,li2025webthinker} interact directly with the live internet, introducing significant training variance and prohibitive costs that limit continuous capability growth. Local retrieval-based systems~\citep{jin2025search,li2025search,sun2025zerosearch} are typically confined to a narrow corpus, making it impossible to replicate the complex search dynamics of the diverse internet. Meanwhile, current real-web graph-based data synthesis methods~\citep{jin2025search,li2025websailor,gao2025beyond} rely on over-engineered reasoning structures, focusing on handcrafted logic while ignoring the diversity of atomic search techniques required for deep research, such as cross-verification and enumeration.

 We argue that scaling agentic RL for Deep Research requires a virtual world that mirrors the structure of the live internet but is isolated in execution: the twin architecture ensures that policies optimized in the virtual environment can generalize to the open web, while isolation shields the training process from the high variance and stochasticity of environmental noise.
 
In this paper, we present \textbf{LiteResearcher}, an Agentic RL training framework designed to construct a \textbf{virtual world that mirrors real-world web architecture}. We follow these steps to realize the scaling potential of agentic RL: (1) Synthesize real search data; (2) Simulate search environment; (3) Achieve sustained improvements through curriculum RL. 

The foundation of this virtual world is generating realistic training tasks. To capture the diversity of information seeking in real-world user queries, we prioritize the scale of information sources in our virtual world. We find that by scaling up information sources, even a \textbf{simple-but-effective} data synthesis can yield a realistic distribution of training tasks. 

Crucially, handling these synthetic tasks requires a corresponding execution environment. We simulate a realistic search environment by enriching our corpus with these tasks. For every synthetic task, we perform a targeted expansion of the local webpage corpus by searching and fetching related real-world webpages, which then construct the local search and browse tool. This ensures that for any given training task, the local search and browse tools provide search dynamics analogous to the open web, allowing the agent to obtain real-web atom capabilities in a controlled environment.

Even with synthetic tasks and tools, training can still saturate if the model gets trapped at a certain difficulty threshold, correctly solving easy tasks while completely failing complex ones. We apply a difficulty-aware curriculum RL which progressively enhances both task difficulty and context length across training stages. At each stage, we only keep the synthetic tasks that the current model can partially solve, discarding both trivially easy and impossibly hard instances, thereby providing consistent training signal for improvement.

Our tiny search agent \textbf{LiteResearcher-4B} achieves 71.3\% on GAIA~\citep{mialon2023gaia} and 78.0\% on Xbench~\citep{chen2025xbench}, surpassing open-source agents up to $8\times$ larger such as Tongyi DeepResearch 30B (70.9\%/75.0\%) and WebSailor 30B (53.2\%/53.3\%), while matching or exceeding proprietary systems including Claude-4.5-Sonnet (71.2\%/66.0\%) and GLM-4.6 (71.9\%/70.0\%). These results demonstrate that scalable RL can yield a powerful on-device deep research agent.

\section{Related Work}

\paragraph{Reinforcement Learning for Reasoning}
The pursuit of advanced reasoning in LLMs initially relied on inference-time heuristics such as Chain-of-Thought prompting~\citep{wei2022chain} and Tree of Thoughts~\citep{yao2023tree}. A paradigm shift occurred with the demonstration that RL can internalize reasoning capabilities directly into model weights. DeepSeek-R1~\citep{guo2025deepseek} showed that RL with verifiable rewards (RLVR) is sufficient to unlock emergent reasoning behaviors such as self-verification and backtracking, and subsequent work has rapidly advanced optimization~\citep{yu2025dapo,liu2025prorl}, reward design~\citep{cui2025process}, and scaling~\citep{hu2025open,yang2025qwen3}, establishing RLVR as a general-purpose post-training paradigm. However, all these methods operate as closed-world reasoners: they rely solely on parametric knowledge and lack the ability to interact with external dynamic environments---a critical limitation for tasks requiring real-time information retrieval.

\paragraph{Agentic Deep Research Models}
Recent research has pivoted towards Agentic Deep Research, where agents interact with external search environments through iterative query formulation and page navigation. This line of work can be grouped by environment design. Online real-world RL systems~\citep{zheng2025deepresearcher,li2025webthinker,li2025websailor} train directly on the live web, offering realistic interactions but incurring high cost and non-deterministic reward signals. Local retrieval-based systems~\citep{jin2025search,li2025search,gao2025beyond} construct environments from Wikipedia or similar corpora, providing fast, deterministic interactions but remaining confined to narrow, homogeneous domains. Simulated systems such as ZeroSearch~\citep{sun2025zerosearch} replace search engines by training an LLM to mimic search engine outputs, but sacrifice page-level fidelity. Recent lightweight explorations such as AgentCPM-Explore~\citep{chen2026agentcpm} further show the capability of small deep-research agents, but still fail to realize scalable RL training. In contrast, our framework builds a local environment with real-world data and a search engine, enabling scalable RL training.

\section{Preliminaries: Agentic Framework}
\label{sec:preliminaries}

We model deep research as a sequential decision-making process under the ReAct framework \citep{yao2022react}. Given an initial user query $q$, the agent interacts with the environment over multiple steps by alternately generating reasoning thoughts, executing actions, and receiving observations. The history up to step $t$ is denoted by
\begin{equation*}
    \mathcal{H}_t = (q, \tau_1, a_1, o_1, \dots, \tau_t, a_t, o_t),
\end{equation*}
where $\tau_i$, $a_i$, and $o_i$ denote the thought, action, and environmental observation at step $i$, respectively. At each step, the agent first generates a reasoning thought $\tau_t$ conditioned on the preceding history, and then selects an executable action $a_t$ conditioned on both $\mathcal{H}_{t-1}$ and $\tau_t$:
\begin{equation*}
    \tau_t \sim \pi_\theta(\cdot \mid \mathcal{H}_{t-1}), \qquad a_t \sim \pi_\theta(\cdot \mid \mathcal{H}_{t-1}, \tau_t).
\end{equation*}

\textbf{Action Space.} The action space $\mathcal{A}$ consists of two primitive actions:
\begin{itemize}
    \item \textbf{Search}$(q')$: submits a natural-language query $q'$ to the search engine and returns a ranked list of retrieved snippets and URLs.
    \item \textbf{Browse}$(u, q')$: visits a selected URL $u$ and returns a query-conditioned summary of the page content.
\end{itemize}
After collecting sufficient information, the agent terminates environment interaction and generates a final response to the original user query $q$.

\section{Methods: Towards Scalable Training}

\begin{figure}[t]
    \centering
    \makebox[\textwidth][c]{\includegraphics[width=\textwidth]{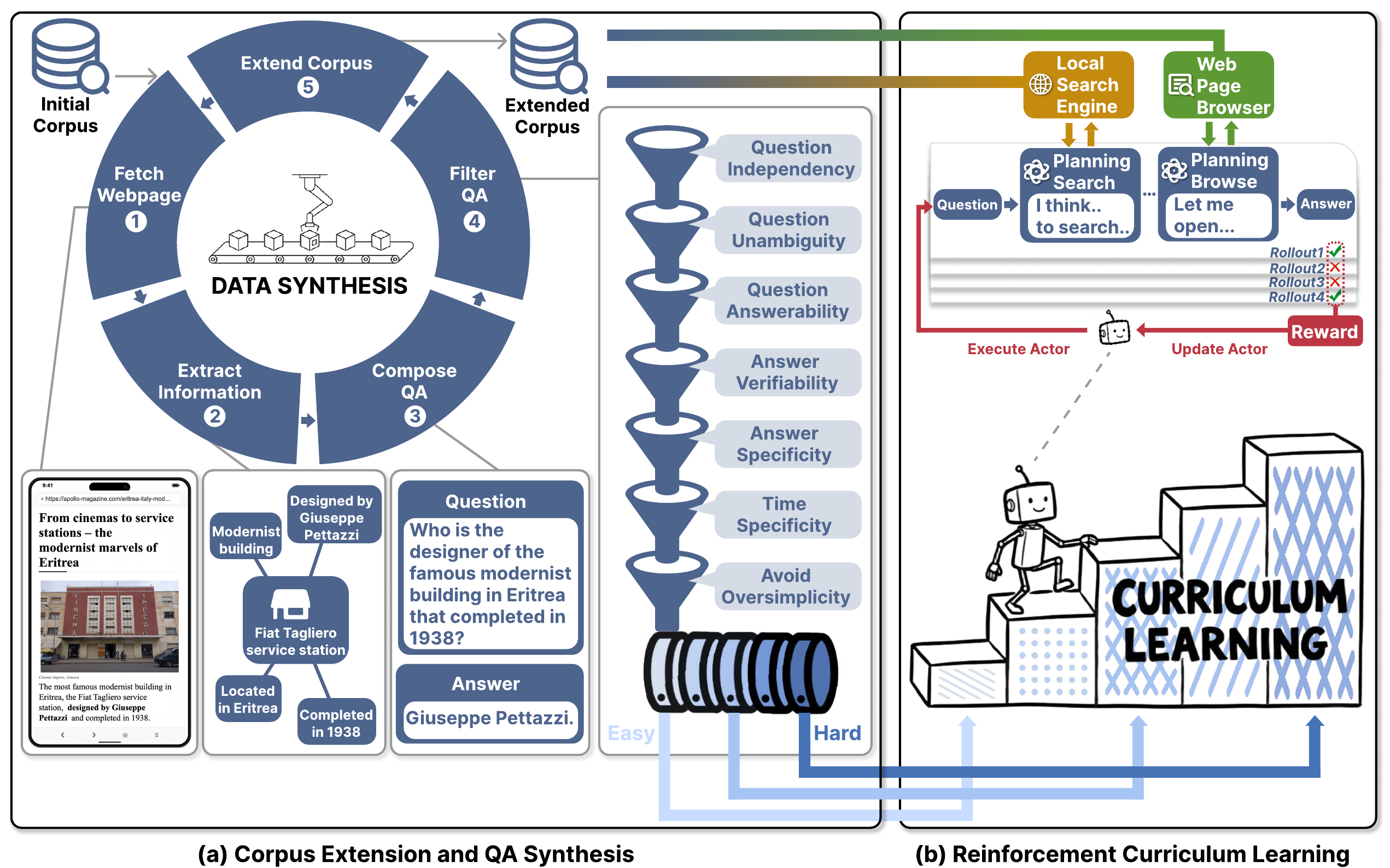}}
    \caption{System architecture overview. (a) \textbf{Corpus Extension and QA Synthesis}: An iterative data engine which also enriches local webpage corpus, powering stable, local tools for zero-cost agent RL training. (b) \textbf{Reinforcement Curriculum Learning}: Synthetic tasks are leveled by complexity to guide the agent through progressive training stages. This reinforcement learning loop utilizes local tool interactions, scaling research capabilities while effectively preventing training saturation.}
    \label{fig:system_architecture}
\end{figure}

% ==========================================
To explore the scaling potential of Agentic Deep Research, we aim to establish a scalable training framework as a mandatory first step. Current paradigms that rely on high-latency, unstable online interactions fundamentally bottleneck this process. In this section, we present our LiteResearcher framework, which decouples agent evolution from the open web while ensuring generalization to real-world web dynamics. As illustrated in Figure~\ref{fig:system_architecture}, the ecosystem is built upon three core pillars:

(1) \textbf{Co-construct Training Data and Corpus Pipeline}: To achieve \textbf{diversity} in the data distribution and \textbf{scalability} in data quantity, we propose a unified framework that co-evolves the training data and the local corpus. 

(2) \textbf{Stable Local Tool Environment}: Leveraging the enriched webpage corpus, we construct a Local Search Engine and Local Browse Tool that mirror the behavior of real-world search dynamics within the synthesized data.

(3) \textbf{Difficulty-Aware Curriculum Learning}: To continuously expand the model's capability boundaries, we leverage our unlimited synthetic data pool to implement a multi-stage curriculum that serves the model with data of appropriate difficulty, thereby ensuring sustainable performance improvements.

\setlength{\textfloatsep}{1.5em}
\begin{table}[t]
    \centering
    \footnotesize
    \renewcommand{\arraystretch}{1.2}
    \setlength{\tabcolsep}{3pt}       

    \begin{tabularx}{\textwidth}{m{1.8cm} l >{\raggedright\arraybackslash}m{5.5cm} >{\raggedright\arraybackslash}X}
        \toprule
        \textbf{Capability} & \textbf{Graph} & \textbf{Example in our synthetic data} & \textbf{Golden Path} \\ \midrule
        
        Direct information & 
        \includegraphics[width=1.5cm, keepaspectratio, valign=c]{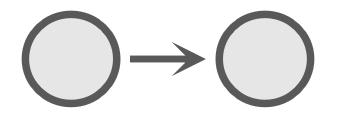} & 
        What is the total global population in 2025? & 
        1. Search for reliable webpage \newline 2. Browse the page to find the value \\ \midrule
        
        Aggregation & 
        \includegraphics[width=1.5cm, keepaspectratio, valign=c]{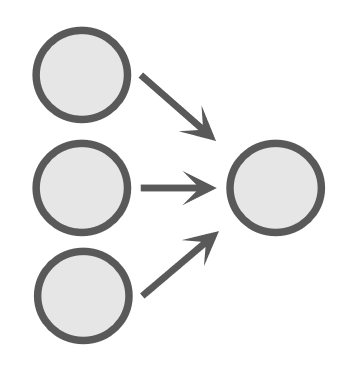} & 
        In the Apollo Belvedere statue restoration project completed in October 2024, what was the material used by Andrea Felice, who employed traditional techniques, to create the replica that was used to replace the missing left hand of the original statue? & 
        1. Summarize all constraints \newline 2. Find solution of each \newline 3. Take the intersection \\ \midrule
        
        Enumeration & 
        \includegraphics[width=1.5cm, keepaspectratio, valign=c]{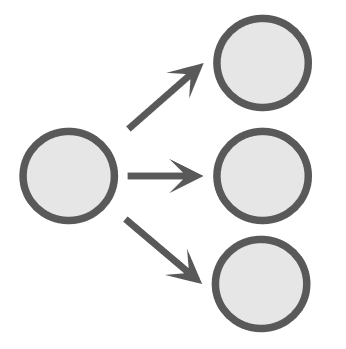} & 
        How many stations were there in total when the southeast section of Edmonton's Valley Line (from the city centre to Mill Woods) opened on November 4, 2023? & 
        1. Brainstorm all possible sources \newline 2. Visit each source separately \newline 3. Take the union set \\ \midrule
        
        Cross verify & 
         \includegraphics[width=1.88cm, keepaspectratio, valign=c]{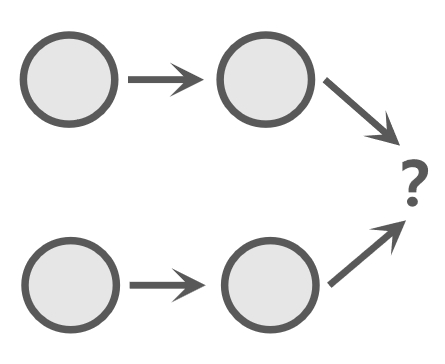} & 
        Does the product ``Quancong High Zinc High Calcium Milk Powder for Students'' under the brand of Inner Mongolia Yili Industrial Group Co., Ltd. contain sucrose? If so, what is the quantity? & 
        1. Find multiple sources \newline 2. Cross verify across sources \\ \midrule
        
        Statistics & 
        \includegraphics[width=1.5cm, keepaspectratio, valign=c]{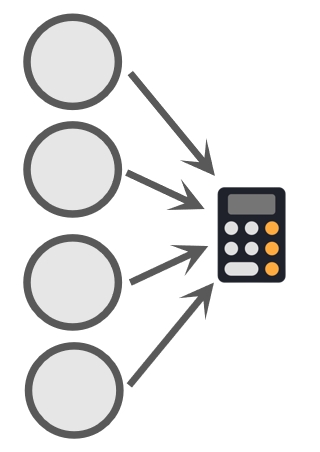} & 
        What was the R-squared (R\textsuperscript{2}) value of the Linear Regression model used to predict future drug-resistant tuberculosis (DR-TB) cases in the Oliver Reginald Tambo District Municipality, South Africa, based on historical data from 2018 to 2021? & 
        1. Find relevant data \newline 2. Calculate R-squared value \\
        \bottomrule
    \end{tabularx}

\caption{Five atomic search capabilities fundamental to complex deep research.}
\label{tab:cases_display}

\end{table}

\subsection{Co-construct Training Data and Corpus Pipeline}
\label{data_gen}
We identify that complex real-world deep research trajectories are fundamentally composed of five atomic search capabilities:
\begin{itemize}[leftmargin=*, itemsep=1ex, parsep=0.5ex]
    \item \textbf{Direct Information}: Directly finding a specific piece of information via search.
    \item \textbf{Aggregation}: Locating a target answer based on multiple identifying attributes.
    \item \textbf{Enumeration}: Listing and counting all entities that satisfy specific constraints.
    \item \textbf{Cross-verification}: Validating and triangulating claims across independent sources.
    \item \textbf{Statistics}: Performing comprehensive calculations and extracting numerical metrics.
\end{itemize}

To construct a dataset with both capability \textbf{diversity} and \textbf{scalability}, rather than relying on complex, handcrafted generation logic, we first scale up the underlying information sources to comprehensively cover the diverse search patterns of the real world. Subsequently, we filter this vast information pool into solvable, high-quality tasks. Specifically:

\paragraph{QA generation from Seed Corpus} 
The data construction process begins with a high-quality \textbf{Seed Corpus} including Wikipedia and BBC News articles. We then prompt an LLM with the template in Appendix~\ref{sec:app_extract_prompt}
 to process this raw content and extract structured question-answer (QA) pairs that target key factual entities within the content. These initial QA pairs serve as the foundational ``seed tasks'' that trigger the subsequent iterative expansion of the local environment.

\paragraph{Information Source Masking} 
To prevent the agent from relying on trivial shortcuts, we implement a \textbf{source masking} mechanism. After a QA pair is generated, we deliberately delete its original information source from the local corpus. This forces the agent to navigate the expanded corpus to discover alternative, non-trivial search pathways, naturally eliciting the complex atomic search capabilities detailed in Table~\ref{tab:cases_display}.

\paragraph{Strict Quality and Difficulty Filtering} 
To guarantee the quality and solvability of the dataset, the candidate QA pairs undergo a rigorous LLM-based evaluation guided by a \textbf{7-point rubric} (Appendix~\ref{sec:app_filter_prompt}). As detailed in Appendix~\ref{sec:app_filter_prompt}, this rubric contains dimensions such as independence, ambiguity, answer verifiability, etc. A data point is retained only if it strictly satisfies all seven criteria. Furthermore, prior to each training session, we apply an additional \textbf{difficulty filter} to ensure the tasks are appropriately calibrated to the agent's current capability (see Section~4.3). This final distillation process provides a \textbf{diverse and realistic data distribution} for subsequent training.

\paragraph{Continuously Expanding Corpus} 
For each filtered QA pair, we utilize a search engine to fetch related webpages from the live internet. These newly retrieved pages are continuously appended to the \textbf{Enriched Corpus}. This iterative expansion creates a corpus covering the vast and diverse information distribution of the real world.

\subsection{Stable Local Tool Environment}

Leveraging the enriched corpus from Section~\ref{data_gen} ($\sim$32M pages, 1M+ domains), we build a fully local tool environment that replaces online interaction during RL training. Unlike standard RAG pipelines that split each page into many chunks, we adopt \textbf{page-level indexing}: each webpage is represented by a single vector (title + summary). This keeps the index small enough to serve hundreds of concurrent rollouts at low latency, which would be infeasible with a chunk-level index that is $\sim$10$\times$ larger. During training, the system sustains hundreds of concurrent tool requests, accumulating to 73.2M calls over the full RL run (Section~\ref{sec:experiments}). The corpus is served through two purpose-built services (infrastructure details in Appendix~\ref{sec:app_infra}):

\begin{itemize}[leftmargin=*,nosep]
    \item \textbf{Local Search Engine}: We adopt BGE-M3~\citep{chen2024bge} for hybrid retrieval, which produces both dense and learned sparse embeddings in a single forward pass. The two signals are fused at query time through the built-in hybrid search of Milvus~\citep{wang2021milvus}. All embeddings are stored on disk and accessed via DiskANN~\citep{jayaram2019diskann} with mmap, achieving $\sim$0.15\,s/query ($\sim$10$\times$ faster than online search engine).
    \item \textbf{Local Browse Tool}: Full webpage content is stored as Markdown in PostgreSQL, keyed by URL and tuned for 1,000 concurrent connections; pages are returned directly upon request at $\sim$0.17\,s/page ($\sim$46$\times$ faster than Jina Reader).
\end{itemize}

\noindent

\subsection{Difficulty-Aware Curriculum Learning}

A fundamental challenge in Agentic RL is training saturation, where the reward curve converges because the model becomes trapped at a difficulty threshold, correctly solving easy tasks while completely failing complex ones, resulting in no meaningful learning gradient.

\paragraph{Difficulty-Aware Data Filtering}
To address training saturation, before each RL stage, we evaluate the difficulty of training queries. For each query, we generate $K=8$ rollouts (pass@8) and strictly retain only those where the number of correct responses $c$ satisfies $1 \le c \le 7$. Queries with $c=8$ are discarded as trivial (yielding zero useful gradient), while those with $c=0$ are deemed impossible or excessively noisy.

\paragraph{On-Policy GRPO Curriculum Learning}
We employ the Group Relative Policy Optimization (GRPO) \citep{shao2024deepseekmath} framework. Given a query $q$ and sampled outputs $\{o_1, \dots, o_K\}$, the advantage $A_i$ is computed by normalizing the reward of $o_i$ against the group mean and standard deviation. Stripped of KL and entropy terms, our objective function is defined strictly as the clipped surrogate loss:
$$
\mathcal{J}_{GRPO}(\theta) = \mathbb{E}_{q \sim P(Q), \{o_i\}_{i=1}^K \sim \pi_{\theta_{old}}} \left[ \frac{1}{K} \sum_{i=1}^K \min \left( r_i(\theta) A_i, \text{clip}\left( r_i(\theta), 1-\epsilon_{low}, 1+\epsilon_{high} \right) A_i \right) \right]
$$
where $r_i(\theta) = \frac{\pi_\theta(o_i|q)}{\pi_{\theta_{rollout}}(o_i|q)}$ is the probability ratio of training engine and rollout engine. While standard implementations perform multiple mini-batch updates over a replay buffer, we enforce a strictly \textit{on-policy} update. We present a detailed ablation study regarding this completely on-policy formulation in Section \ref{sec:experiments}.

\section{Experiments}
\label{sec:experiments}
\subsection{Experimental Setup}

We evaluate \textbf{LiteResearcher-4B} on a suite of complex reasoning and agentic search benchmarks,
including GAIA~\citep{mialon2023gaia}, BrowseComp~\citep{wei2025browsecomp}, HLE~\citep{phan2025hle},
Frames~\citep{krishna2024frames}, WebWalker~\citep{wu2025webwalker}, Seal-0~\citep{pham2025sealqa},
and Xbench-DeepSearch-2505~\citep{chen2025xbench}. For BrowseComp, we randomly sampled 400 examples for evaluation. Evaluation is conducted with online APIs (Serper for search and Jina for browsing), following the same tool setup as prior work~\citep{team2025tongyi}. We compare against leading closed-source foundation models, representative open-source deep-research agents, and AgentCPM-Explore~\citep{chen2026agentcpm} as a concurrent lightweight baseline. 

All experiments are initialized from Qwen3-4B-Thinking-2507~\citep{yang2025qwen3}. We first perform a cold-start SFT stage to teach the model basic tool-use capabilities such as query formulation, search invocation, and page browsing (details in Appendix~\ref{sec:app_sft}). The RL stage generates rollouts entirely within our local environment, with a global batch size of 128 (on-policy update), generating 8 rollouts per query. The learning rate is fixed at $1 \times 10^{-6}$. The maximum response length is initialized at 32K tokens in the first stage and expanded by 16K tokens in each subsequent stage. Our loss function uses neither a KL divergence penalty nor an entropy regularization term. 

\subsection{Main Results}

\begin{table}[h]
    \centering
    \footnotesize 
    \setlength{\tabcolsep}{4pt} 
    \renewcommand{\arraystretch}{1.1} 
    \begin{tabular}{l ccc ccc cc}
        \toprule
        \textbf{Models} & 
        \begin{tabular}{@{}c@{}}\textbf{GAIA-} \\ \textbf{Text}\end{tabular} & 
        \begin{tabular}{@{}c@{}}\textbf{Browse-} \\ \textbf{comp}\end{tabular} & 
        \begin{tabular}{@{}c@{}}\textbf{Browse.} \\ \textbf{(ZH)}\end{tabular} & 
        \textbf{HLE} & 
        \textbf{Frames} & 
        \begin{tabular}{@{}c@{}}\textbf{Web-} \\ \textbf{walker}\end{tabular} & 
        \textbf{Seal-0} & 
        \begin{tabular}{@{}c@{}}\textbf{Xbench-} \\ \textbf{DS}\end{tabular} \\
        \midrule
        \multicolumn{9}{l}{\textit{\textbf{Commercial Models}}} \\
        Claude-4-Sonnet & 68.3\% & 12.2\% & 29.1\% & 20.3\% & 80.7\% & 61.7\% & - & 64.6\% \\
        Claude-4.5-Sonnet & 71.2\% & 19.6\% & 40.8\% & 24.5\% & 85.0\% & - & 53.4\% & 66.0\% \\
        DeepSeek-V3.2 & 63.5\% & 67.6\% & 65.0\% & 40.8\% & 80.2\% & - & 38.5\% & 71.0\% \\
        DeepSeek-V3.1 & 63.1\% & 30.0\% & 49.2\% & 29.8\% & 83.7\% & 61.2\% & - & 71.0\% \\
        Minimax-M2 & 75.7\% & 44.0\% & 48.5\% & 31.8\% & - & - & - & 72.0\% \\
        OpenAI-GPT-5-high & 76.4\% & 54.9\% & 65.0\% & 35.2\% & - & - & 51.4\% & 77.8\% \\
        GLM-4.6 & 71.9\% & 45.1\% & 49.5\% & 30.4\% & - & - & - & 70.0\% \\
        Kimi-Researcher & - & - & - & 26.9\% & 78.8\% & - & 36.0\% & 69.0\% \\
        Kimi-K2-0905 & 60.2\% & 7.4\% & 22.2\% & 21.7\% & 58.1\% & - & 25.2\% & 61.0\% \\
        \midrule
        \multicolumn{9}{l}{\textit{\textbf{Open-Source Models}}} \\
        Mirothinker 8B & 66.4\% & 31.1\% & 40.2\% & 21.5\% & 80.6\% & 60.6\% & 40.4\% & 60.6\% \\
        Tongyi DeepResearch 30B & 70.9\% & \textbf{43.4\%} & \textbf{46.7\%} & \textbf{32.9\%} & \textbf{90.6\%} & 72.2\% & - & 75.0\% \\
        ASearcher QWQ v2 & 58.7\% & - & - & - & 74.5\% & - & - & 51.1\% \\
        WebSailor 30B & 53.2\% & - & - & - & - & - & - & 53.3\% \\
        WebDancer (QwQ) & 51.5\% & 3.8\% & 18.0\% & - & - & 47.9\% & - & 38.3\% \\
        WebExplorer-8B & 50.0\% & 15.7\% & 32.0\% & 17.3\% & 75.7\% & 62.7\% & - & 53.7\% \\
        DeepMiner-32B & 58.7\% & 33.5\% & 40.1\% & - & - & - & - & 62.0\% \\
        AFM-RL-32B & 55.3\% & 11.1\% & - & 18.0\% & - & 63.0\% & - & - \\
        SFR-DeepResearch & 66.0\% & - & - & 28.7\% & 82.8\% & - & - & - \\
        AgentCPM-Explore-4B & 63.9\% & 24.1\% & 29.1\% & 19.1\% & 82.7\% & 68.1\% & 40.5\% & 70.0\% \\
        \midrule
        \rowcolor{blue!10} \textbf{LiteResearcher-4B} & \textbf{71.3\%} & 27.5\%* & 32.5\%* & 22.0\% & 83.1\% & \textbf{72.7\%} & \textbf{41.8\%} & \textbf{78.0\%} \\
        \bottomrule
    \end{tabular}
    \caption{\textbf{Main results.} Comparison across commercial models and open-source deep research agents on eight benchmarks. Best open-source results are in \textbf{bold}. Results without * are evaluated with a 128k context window; results with * use a 64k context window with a memory mechanism that, upon reaching the context limit, calls a summarization model to compress each previous tool interaction step into one sentence.}
    \label{tab:main_results_1}
\end{table}
\nocite{anthropic2025claude4_5,liu2024deepseekv3,minimax2025m2,openai2025gpt5,zeng2025glm,moonshot2025kimi_researcher,team2025kimi,team2025mirothinker,team2025tongyi,li2025websailor,wu2025webdancer,liu2025webexplorer,tang2025beyond,li2025chain,nguyen2025sfr,chen2026agentcpm}

Table~\ref{tab:main_results_1} summarizes the performance of \textbf{LiteResearcher (4B)} across eight benchmarks spanning general-purpose agentic search, multi-hop reasoning, and cross-lingual deep research. From the results, we make the following key observations.

\textbf{A 4B model trained locally approaches commercial-level performance.} LiteResearcher achieves \textbf{71.3\%} on GAIA-Text, matching Claude-4.5-Sonnet (71.2\%) and surpassing Tongyi DeepResearch 30B (70.9\%) and other open-source models with 8--32B parameters. On Frames, it scores 83.1\%, outperforming Claude-4-Sonnet (80.7\%) and DeepSeek-V3.2 (80.2\%). On Xbench-DS, a comprehensive bilingual benchmark, LiteResearcher reaches 78.0\%, the highest among all open-source models and surpassing OpenAI-GPT-5-high (77.8\%). These results suggest that the data--environment bottleneck is at least as important as model scale for strong agentic search performance.

\textbf{Local training matches or exceeds online training.} AgentCPM-Explore-4B~\citep{chen2026agentcpm}, a concurrent 4B model trained on the live internet, is outperformed by LiteResearcher on all 8 benchmarks (e.g., GAIA 71.3\% vs.\ 63.9\%, Xbench 78.0\% vs.\ 70.0\%). AgentCPM-Explore further reports that online environment instability is a major source of reward noise, limiting its RL gains to only +3.8\% over the SFT baseline on GAIA. Our deterministic local environment eliminates such noise, sustaining over 700 stable RL steps with monotonic improvements.

\textbf{Context length and BrowseComp.} BrowseComp requires exceptionally deep browsing chains, often exceeding 20 pages per query. For a 4B model, this frequently exhausts the 128K context window, causing performance degradation (20.3\%/28.4\% without memory, Table~\ref{tab:sft_baseline}). A memory mechanism that summarizes prior tool interactions at 64K mitigates this, improving results to 27.5\%/32.5\%. This gap highlights long-context utilization as a key limitation of small-scale agents, and we report both configurations for transparency.

\subsection{Analysis and Ablations}
\label{sec:analysis}

\paragraph{Local Environment vs.\ Live Internet.}
Over the full RL run, our agents issued 45.8M search queries and 27.4M browse requests---totaling \textbf{73.2M tool calls}. As shown in Table~\ref{tab:local_vs_online}, fulfilling this volume through commercial search engines and browser proxies would cost \$59K--\$243K, whereas our fully local pipeline incurs zero marginal cost. Beyond the cost saving, the 10--46$\times$ latency advantage directly translates to higher rollout throughput, making our local ecosystem a prerequisite---not merely an optimization---for scalable on-policy agentic RL.

\begin{table}[h]
    \centering
    \small
    \setlength{\tabcolsep}{5pt}
    \renewcommand{\arraystretch}{1.1}
    \begin{tabular}{ll rrrrr}
        \toprule
        \textbf{Tool} & \textbf{Provider} & \textbf{Latency} & \textbf{Speedup} & \textbf{Price/K} & \textbf{\#Calls} & \textbf{Est.\ Cost} \\
        \midrule
        Search & Serper$^{a}$ & $\sim$1.5\,s & -- & \$1 & 45.8M & \$45,802 \\
        Search & SerpAPI$^{b}$ & $\sim$2\,s & -- & \$5 & 45.8M & \$229,012 \\
        Browse & Jina Reader$^{c}$ & 7.9\,s & -- & $\sim$\$0.50 & 27.4M & \$13,709 \\
        \midrule
        \multicolumn{2}{l}{Online total} & -- & -- & -- & 73.2M & \$59K--\$243K \\
        \rowcolor{blue!8}
        \multicolumn{2}{l}{\textbf{Ours (fully local)}} & \textbf{0.15 / 0.17\,s} & \textbf{10--46$\times$} & \textbf{\$0} & \textbf{73.2M} & \textbf{\$0} \\
        \bottomrule
    \end{tabular}
    \caption{\textbf{Local vs.\ Online}: latency and cost for all 73.2M tool calls during RL. $^{a}$\href{https://serper.dev}{Serper}. $^{b}$\href{https://serpapi.com}{SerpAPI}. $^{c}$\href{https://jina.ai/reader}{Jina Reader}, token pricing at \$0.05/M tokens ($\sim$10K tokens/page).}
    \label{tab:local_vs_online}
\end{table}

\noindent
\begin{minipage}[t]{0.55\textwidth}
    \paragraph{Effect of Our Synthetic Data.}
    Our synthesis pipeline produces data diverse enough to cover real-world search patterns and support strong generalization (Table~\ref{tab:ablation_data}): combining our synthetic data with multi-hop data achieves 66.8\% on GAIA and 71.0\% on Xbench. Removing our synthetic data drops performance to 58.7\%/66.3\%, confirming that it covers diverse search patterns.
\end{minipage}
\hfill
\begin{minipage}[t]{0.4\textwidth}
    \vspace{0.3em}
    \centering
    \small
    \setlength{\tabcolsep}{6pt}
    \begin{tabular}{lcc}
        \toprule
        \textbf{Training Data} & \textbf{GAIA} & \textbf{Xbench} \\
        \midrule
        Multi-hop only & 58.7\% & 66.3\% \\
        Ours + multi-hop & \textbf{66.8\%} & \textbf{71.0\%} \\
        \bottomrule
    \end{tabular}
    \captionof{table}{\textbf{Effect of our synthetic data.}}
    \label{tab:ablation_data}
\end{minipage}

\vspace{-0.7em} 
\begin{figure}[htbp]
    \centering
    \includegraphics[width=\textwidth]{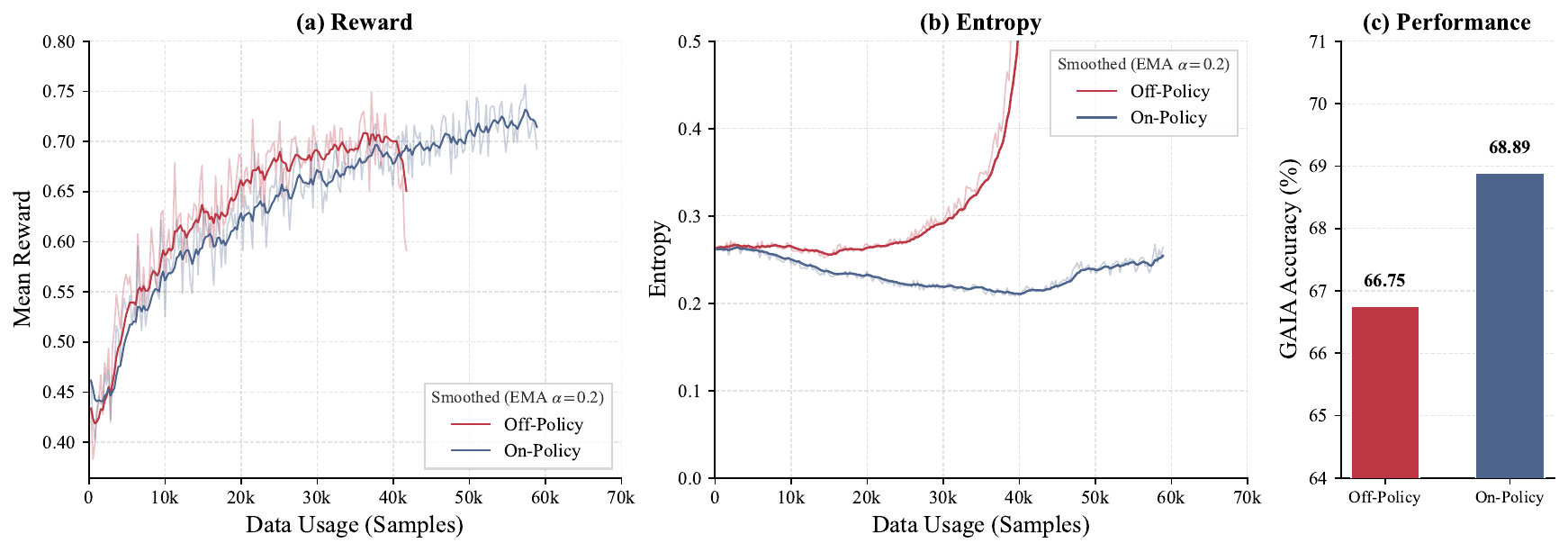}
    \caption{\textbf{On-Policy vs.\ Off-Policy training reward.} On-policy training is more stable and continues to improve throughout training.}
    \label{fig:ablation_curves}
\end{figure}
\paragraph{On-Policy vs.\ Off-Policy Training.}
We compare a strictly on-policy algorithm, where each rollout batch is consumed in a single update and then discarded, with a more off-policy algorithm, where each rollout batch is split into multiple mini-batches (e.g., 256 samples into 4 mini-batches) and used for several successive updates. As shown in Figure~\ref{fig:ablation_curves}, the off-policy variant gains reward much earlier but eventually declines, whereas on-policy training continues to improve more steadily throughout training. Figure~\ref{fig:ablation_curves} shows that on-policy training achieves higher GAIA validation accuracy (68.9\%) than off-policy training (66.8\%).

These results suggest that long-horizon Agentic RL is particularly sensitive to off-policy updates. We hypothesize that multiple updates on the same rollout batch introduce policy lag, leading to increasing mismatch between the sampled trajectories and the current policy. In long-horizon search tasks, such mismatch can accumulate along the trajectory, harming optimization stability and ultimately limiting the achievable performance ceiling.

\begin{figure}[htbp]
    \centering
    \includegraphics[width=0.7\textwidth]{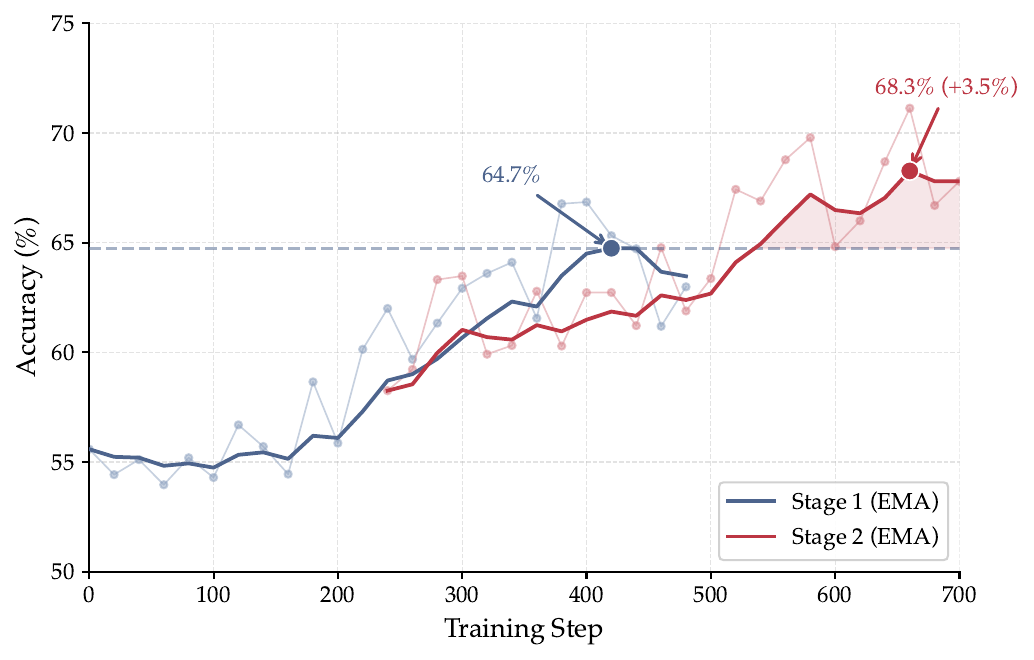}
    \caption{\textbf{Stage 1 vs.\ Stage 2.} GAIA accuracy (EMA smoothed) during RL training. The two-stage curriculum overcomes the Stage~1 plateau.}
    \label{fig:stage_ablation}
\end{figure}

\paragraph{Effectiveness of Multi-Stage Training.}
Single-stage RL saturates once the model exhausts tasks at its current difficulty level; deliberate difficulty escalation is the key to continued improvement. Figure~\ref{fig:stage_ablation} shows the learning curves of our two-stage strategy. In Stage~1, the model reaches a plateau at 64.7\% GAIA accuracy. After transitioning to Stage~2 with adjusted data distribution and targeted complexity, training continues to improve, reaching 68.3\% (+3.6\%). This indicates that the second stage helps sustain learning beyond the saturation point of the first, enabled by our unlimited synthetic data pool.

\begin{figure}[t]
    \centering
    \includegraphics[width=\textwidth]{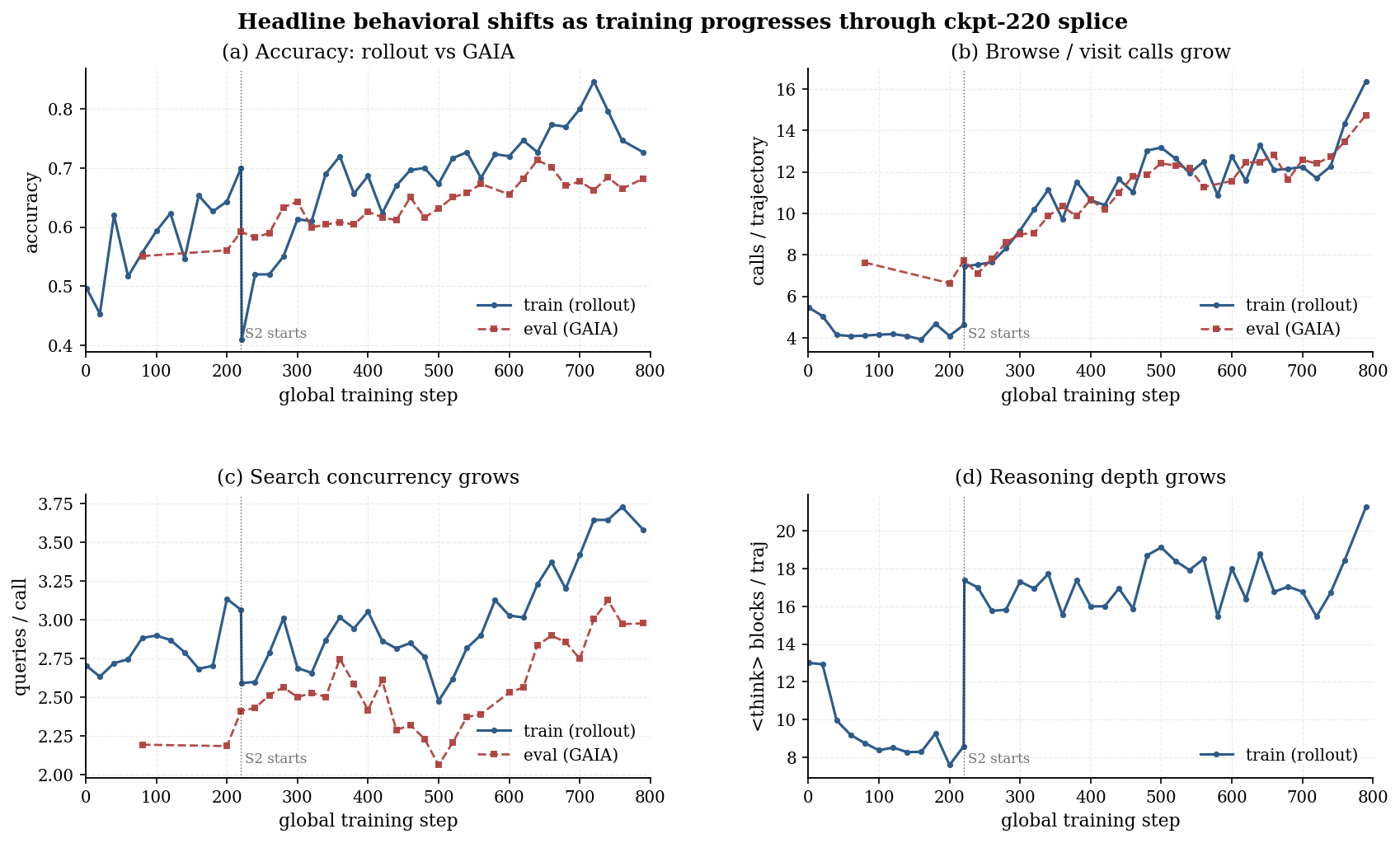}
    \caption{\textbf{Behavior evolution along the full training trajectory} (Stage~1 steps~0--220 continued by Stage~2 at the ckpt-220 splice). Training rollouts (solid blue) and held-out GAIA evaluations (dashed red) share a global-step axis: \textbf{(a)}~accuracy, \textbf{(b)}~browse/visit calls per trajectory, \textbf{(c)}~search concurrency (queries per call), and \textbf{(d)}~reasoning depth (\texttt{<think>} blocks per trajectory). All four behaviors grow as training progresses, and the rollout and evaluation curves move in lock-step---evidence of transferable rather than overfit behavior.}
    \label{fig:behavior_evolution}
\end{figure}

\paragraph{Behavior Evolution Along Training.}
Beyond aggregate accuracy, we examine how the agent's \emph{behavior} evolves over the entire training run, placing Stage~1 (steps~0--220) and Stage~2 on a single continuous global-step axis, since Stage~2 warm-starts from the Stage~1 step-220 checkpoint. For each metric we plot both the on-policy training rollouts and the held-out GAIA evaluations. Figure~\ref{fig:behavior_evolution} reveals four coupled behavioral shifts. \textbf{(a)}~Training and GAIA accuracy climb together throughout, reaching GAIA pass@1 $\approx$\,0.68. \textbf{(b)}~Browse/visit calls per trajectory grow near-linearly across Stage~2: the agent increasingly \emph{reads} pages rather than only issuing searches. \textbf{(c)}~Search concurrency (queries per call) rises, indicating that the agent learns to batch parallel queries instead of probing serially. \textbf{(d)}~Reasoning depth, measured by \texttt{<think>} blocks per trajectory, grows steadily until long-horizon planning becomes the dominant mode. Crucially, for every jointly measurable behavior the training-rollout and GAIA-evaluation curves move \emph{in lock-step}---these are transferable capabilities the policy genuinely acquires, not artifacts of overfitting to the training environment.

\paragraph{Contribution of RL over SFT.}
The SFT stage uses trajectories generated by Tongyi DeepResearch as the teacher model (Appendix~\ref{sec:app_sft}). As shown in Table~\ref{tab:sft_baseline}, SFT alone yields 55.6\% on GAIA---still 15.3 points below the teacher's own 70.9\%. The RL stage not only closes this gap but surpasses it, reaching 71.3\% (+15.7 over SFT), with consistent gains across all eight benchmarks. This indicates that the primary performance driver is the RL training framework rather than teacher distillation.

\paragraph{Case study: needle-in-a-haystack identification on BrowseComp.}
To make the learned behavior concrete, we walk through one rollout on BrowseComp---the hardest benchmark we evaluate, where the answer is pinned down only by many scattered constraints. The agent must name an author from \emph{six} independent biographical clues, \emph{no single web page of which contains the answer}. Across \textbf{140 interaction steps} (70 reasoning turns, 35 searches, 47 page visits) it decomposes the query and systematically rules out 20+ candidate authors on the common clues; when these are exhausted, it \emph{pivots} to the single rarest clue---a childhood nightmare about a wolf---which it reasons ``must appear in an interview,'' searches that detail directly, and locks onto \textbf{Melissa Marr}. It then cross-verifies all six clues against six \emph{independent} sources (Table~\ref{tab:browsecomp_clues}). This is genuine multi-hop synthesis rather than a lookup: the answer exists only as the intersection of evidence the agent itself assembled. The full step-by-step rollout is shown below.

\begin{trajbox}{BrowseComp: identifying an author from six scattered biographical clues (140 steps, judged correct)}
\textbf{Question.}\ As of December 31, 2023, name the author who (i)~was a university instructor, (ii)~whose books have been translated into 20+ languages, (iii)~has worked as an editor, (iv)~whose parents were members of the NRA, (v)~had a recurring nightmare about a wolf creature, and (vi)~has three children, two of which are adopted.

\smallskip
\textbf{Reference:}\ Melissa Marr \hfill \textbf{Model answer:}\ \textbf{Melissa Marr}~\textcolor{trajsearch}{\faCheck}
\smallskip

\trturn{1}
\trquote{Six independent constraints (i)--(vi). The NRA detail (iv) suggests an American author, and the recurring wolf nightmare (v) is the rarest, most distinctive clue --- a good anchor. Let me first try generic combinations of the more common clues.}
\trsearch{\texttt{"parents were members of the NRA" author}; \texttt{"three children" "two adopted" author}; \texttt{"university instructor" "editor" "books have been translated"}}
\trobs{\textnormal{Generic results; no single author satisfies all six clues.}}

\trturn{9}
\trvisit{\texttt{en.wikipedia.org/wiki/Neal\_Knox}}
\trquote{Neal Knox keeps surfacing for the NRA clue, but he is a gun-rights activist who wrote about firearms --- his books were not translated into many languages and he is no novelist. Rule him out.}

\trturn{15}
\trvisit{\texttt{hbr.org/...\,carmine-gallo}; \texttt{en.wikipedia.org/wiki/Carmine\_Gallo}}
\trquote{Carmine Gallo is a Harvard instructor with books in 20+ languages --- promising on (i) and (ii). But nothing ties his parents to the NRA, and the wolf and adoption details don't fit. Not him.}

\trturn{43}
\trvisit{\texttt{en.wikipedia.org/wiki/Stephen\_King}}
\trquote{Stephen King fits ``translated into many languages'', but his father was a traveling vacuum salesman, not an NRA member. Fails clue (iv).}

\trturn{47--51}
\trvisit{\texttt{en.wikipedia.org/wiki/J.K.\_Rowling}; \texttt{.../David\_Baldacci}}
\trquote{J.K.\ Rowling --- not a university instructor, parents not NRA. David Baldacci has 45+ translations but only two children, not three. Both fail.}

\trturn{55--57}
\trvisit{\texttt{en.wikipedia.org/wiki/Jenny\_Sampirisi}; \texttt{amberalogan.com}}
\trquote{Jenny Sampirisi is a novelist, editor \emph{and} university instructor --- but her books aren't in 20+ languages. Amber A.\ Logan has two children, one adopted, not three/two. Close, but no.}

\trturn{101}
\trvisit{\texttt{en.wikipedia.org/wiki/Sheila\_Heti}}
\trquote{Sheila Heti's parents are Hungarian-Jewish immigrants, not NRA members. After 20+ rejected candidates, generic clue combinations are exhausted.}

\trturn{113 (pivot)}
\trquote{New strategy: stop guessing on the common clues. The wolf nightmare is bizarre enough that it must appear in an \emph{interview} --- search that detail directly.}
\trsearch{\texttt{"recurring nightmare" "wolf" "interview" author}}
\trvisit{\texttt{goodreads.com/interviews/...\,Melissa\_Marr}}
\trquote{Great! This interview says: ``I had a recurring nightmare when I was a child about a wolf that used to come in and had long witch-like fingernails.'' That is clue (v) verbatim --- the author is \textbf{Melissa Marr}. Now verify the other five.}

\trturn{119}
\trvisit{\texttt{en.wikipedia.org/wiki/Melissa\_Marr}}
\trquote{Wikipedia: Melissa Marr is a ``former university English teacher'' --- clue (i). And her books are translated into \textbf{28} languages --- clue (ii).}
\trobs{\textnormal{(i) and (ii) confirmed.}~\textcolor{trajsearch}{\faCheck}}

\trturn{127--129}
\trsearch{\texttt{"Melissa Marr" NRA}}
\trvisit{\texttt{lightspeedmagazine.com/...\,melissa-marr}}
\trquote{The Lightspeed interview states she ``grew up in rural Pennsylvania with a father who collected guns and was an NRA member'' --- clue (iv) confirmed.}

\trturn{131--138}
\trsearch{\texttt{Melissa Marr adopted children}; \texttt{Melissa Marr editor anthology}}
\trobs{\textnormal{Editor of anthologies, co-edited with Kelley Armstrong (iii); three children, two adopted (vi).}~\textcolor{trajsearch}{\faCheck}}

\trturn{140}
\trthink{All six independent clues now anchor to the \emph{same} person across multiple distinct sources --- no single page ever contained the answer.}
\transwer{\textbf{Melissa Marr} --- every clue independently verified (Table~\ref{tab:browsecomp_clues}).}
\end{trajbox}

\section{Conclusions}

In this paper, we presented \textbf{LiteResearcher}, an agentic RL training framework that resolves the critical bottlenecks of data diversity and environment stability by constructing an isolated virtual world. Crucially, our work yields three key insights for the future of deep research agents. First, we demonstrate that \textbf{Agentic RL can drive sustained capability improvement}. The training saturation observed in prior works is due to reliance on high-variance live environments or a narrow local corpus, rather than a limitation of the RL paradigm itself. Second, we reveal that a \textbf{simple but scalable synthetic data pipeline}, driven by the continuous expansion of real-world information sources, is significantly more effective than over-engineered, handcrafted reasoning structures. Finally, trained entirely locally at zero marginal cost, our compact LiteResearcher-4B achieves 71.3\% on GAIA and 78.0\% on Xbench-DS. By matching or exceeding commercial models, our results suggest that \textbf{scalable RL training, beyond model scale alone, is a critical enabler for strong deep research capabilities}. The entire training framework, including data synthesis pipelines, local environment infrastructure, and RL training code, will be open-sourced to facilitate future research.

\bibliography{main}
\bibliographystyle{main}

\newpage
\appendix

\section{Synthetic Data}
\label{sec:app_synth}

\subsection{Extract Information and Compose QA Prompt}
\label{sec:app_extract_prompt}

This prompt converts a raw webpage into factual question--answer pairs.
We use it to produce single-hop QA items with short, verifiable answers.

\begin{promptbox}{Extract Information and Compose QA Prompt}
You are a data extraction and Q\&A generation expert. Analyze the following webpage content in markdown format and extract all specific, factual data points.

For each concrete data point you find (numbers, statistics, dates, names, amounts, percentages, etc.), create a high-quality question--answer pair.

\textbf{Requirements:}
\begin{enumerate}
    \item Questions should be specific and reference the exact context from the page.
    \item Answers should be concise and factual (e.g., ``500 billion'', ``2025'', ``15\%'').
    \item Focus on extractable, verifiable data points---NOT opinions or general statements.
    \item Each Q\&A pair should be independently understandable with proper context in the question.
\end{enumerate}

\textbf{Example of a good Q\&A pair:}
\begin{quote}\small
Q: According to the India Skills Report 2025 (Key Findings and Government Initiatives), what is the projected global economic contribution of hybrid work models and digital nomadism by 2030?\\
A: \$500 billion
\end{quote}

Now analyze this webpage and generate Q\&A pairs:
\texttt{\{markdown\_content\}}

\textbf{Output format:} Return a JSON object with this structure:
\begin{lstlisting}[basicstyle=\ttfamily\footnotesize,breaklines=true]
{
  "qa_pairs": [
    {"question": "...", "answer": "..."}
  ],
  "total_pairs": <number>
}
\end{lstlisting}
\end{promptbox}

\subsection{Rubric filter Prompt}
\label{sec:app_filter_prompt}

This prompt checks whether a QA pair is self-contained and objectively answerable.
We use it to filter out vague, open-ended, or non-verifiable items before training.

\begin{promptbox}{Data Quality Filtering Prompt}
Please evaluate the quality of the following question--answer (QA) pair. The QA pair must satisfy \emph{all} of the following conditions to pass.

\textbf{Conditions:}
\begin{enumerate}
    \item \textbf{Question Independence:} The question can be understood without any context.
    \item \textbf{Answer Specificity \& Verifiability:} The answer must be verifiable and consist of specific, concrete information such as:
    \begin{itemize}
        \item A specific number (e.g., ``42'', ``1,500 members'').
        \item A particular name (e.g., ``John Smith'', ``Eiffel Tower'').
        \item An exact model/designation (e.g., ``Model T'', ``Boeing 747'').
        \item A precise date/year (e.g., ``1945'', ``March 15, 1990'').
        \item A specific location (e.g., ``Paris'', ``123 Main Street'').
    \end{itemize}
    The answer should \emph{not} be a descriptive list, explanation, or multiple unverifiable items.
    \item \textbf{Question Unambiguity:} The question must be unambiguous with only one clear interpretation.
    \item \textbf{Question Answerability:} The question is clearly stated and it's obvious what is being asked.
    \item \textbf{Avoid Open-ended Questions:} Must allow for a definite, measurable short answer (not an explanatory response). Questions should \emph{not} use words like ``how'', ``why'', ``\zh{怎么}'', ``\zh{如何}'', ``\zh{为什么}''.
    \item \textbf{Avoid Oversimplicity:} The question and answer must not be common sense; if it's too simple, filter it out.
    \item \textbf{Time Specificity:} If the question refers to time-dependent information, it must include a specific, concrete time constraint. Vague temporal references like ``latest'', ``most recent'', ``as of now'', ``currently'', ``so far'', ``up to now'', ``\zh{到目前为止}'', ``\zh{最新}'' are \emph{not} allowed.
\end{enumerate}

\textbf{QA pair to evaluate:}
\begin{lstlisting}[basicstyle=\ttfamily\footnotesize,breaklines=true]
Question: {question}
Answer:   {answer}
\end{lstlisting}

Please provide your reasoning for each condition, then give your final answer in a box. Answer true if and only if the QA pair satisfies all 7 conditions above. Otherwise, answer false.

\textbf{Examples:}

\noindent\hrulefill

\textbf{Example 1}

\textbf{Question:} Flula Borg appeared in Season 2 Episode 3 ``Like a Boss'' of ``Younger'' on Apple TV in 2024. What important event were Liza and Kelsey preparing for in this episode?

\textbf{Answer:} In ``Younger'' Season 2 Episode 3 ``Like a Boss'', Liza and Kelsey were preparing for the launch of their new publishing imprint while facing massive online criticism.

\textbf{Reasoning:}
\begin{enumerate}[label=\arabic*.]
    \item Question Independence: true - The question is self-contained.
    \item Answer Specificity: false - The answer is descriptive, not a specific fact.
    \item Question Unambiguity: true - Clear what is being asked.
    \item Question Answerability: true - The question is clearly stated.
    \item Avoid Open-ended Questions: false - Requires an explanatory description.
    \item Avoid Oversimplicity: true - Not common sense; requires searching.
    \item Time Specificity: true - Contains specific year ``2024''.
\end{enumerate}
There are multiple false conditions, therefore the answer is \texttt{\textbackslash boxed\{false\}}.

\noindent\hrulefill

\textbf{Example 2}

\textbf{Question:} What was the gun configuration of the first production variant J22A (or J22 UBv) of the J22 fighter aircraft developed by the Swedish Royal Air Administration Aircraft Factory (FFVS) for the Swedish Air Force in 1940?

\textbf{Answer:} 2$\times$8mm machine guns and 2$\times$13.2mm machine guns

\textbf{Reasoning:}
\begin{enumerate}[label=\arabic*.]
    \item Question Independence: true - Fully self-contained.
    \item Answer Specificity: true - Specific technical specification.
    \item Question Unambiguity: true - Clearly asks about a specific configuration.
    \item Question Answerability: true - Clear and answerable.
    \item Avoid Open-ended Questions: true - Requires a precise technical answer.
    \item Avoid Oversimplicity: true - Requires searching historical knowledge.
    \item Time Specificity: true - Contains specific year ``1940''.
\end{enumerate}
All conditions are true, therefore the answer is \texttt{\textbackslash boxed\{true\}}.

\noindent\hrulefill

\textbf{Example 3}

\textbf{Question:} What was one of the main German fighter aircraft models that the Swedish J22 fighter faced during its service in the 1940s?

\textbf{Answer:} FW 190

\textbf{Reasoning:}
\begin{enumerate}[label=\arabic*.]
    \item Question Independence: true - Self-contained question.
    \item Answer Specificity: true - Specific aircraft model.
    \item Question Unambiguity: false - ``one of'' implies multiple correct answers.
    \item Question Answerability: true - Clear what is being asked.
    \item Avoid Open-ended Questions: true - Allows for a specific model name.
    \item Avoid Oversimplicity: true - Requires searching historical knowledge.
    \item Time Specificity: true - Contains specific decade ``1940s''.
\end{enumerate}
There is a false condition, therefore the answer is \texttt{\textbackslash boxed\{false\}}.

\noindent\hrulefill

\textbf{Example 4}

\textbf{Question:} In what year was the latest version of the annual report template file for the Cooperative Innovation High School (CIHS) in North Carolina released?

\textbf{Answer:} 2025

\textbf{Reasoning:}
\begin{enumerate}[label=\arabic*.]
    \item Question Independence: true - Self-contained question.
    \item Answer Specificity: true - Specific year.
    \item Question Unambiguity: true - Clear what is being asked.
    \item Question Answerability: true - The question is clearly stated.
    \item Avoid Open-ended Questions: true - Requires a specific year.
    \item Avoid Oversimplicity: true - Requires searching about CIHS.
    \item Time Specificity: false - ``latest'' is vague and changes over time.
\end{enumerate}
There is a false condition, therefore the answer is \texttt{\textbackslash boxed\{false\}}.

\noindent\hrulefill

\textbf{Example 5}

\textbf{Question:} In June 1937, in a collective school in Valencia, Spain, how did a teacher who had studied at Barcelona's ``Nature School'' (La Farigola) use the natural environment in teaching?

\textbf{Answer:} Organized students to visit vegetable gardens and orange trees, observe plant growth and draw what they saw, using nature as a learning object

\textbf{Reasoning:}
\begin{enumerate}[label=\arabic*.]
    \item Question Independence: true - Self-contained question.
    \item Answer Specificity: false - Descriptive explanation rather than a specific fact.
    \item Question Unambiguity: true - Clear what is being asked.
    \item Question Answerability: true - The question is clearly stated.
    \item Avoid Open-ended Questions: false - Uses ``how'' which invites an explanatory response.
    \item Avoid Oversimplicity: true - Requires searching.
    \item Time Specificity: true - Contains specific time ``June 1937''.
\end{enumerate}
There are multiple false conditions, therefore the answer is \texttt{\textbackslash boxed\{false\}}.

\noindent\hrulefill

\textbf{Example 6}

\textbf{Question:} What was Frances Tiafoe's career tour-level finals record after his loss in the final of the 2025 Houston Men's Clay Court Championship?

\textbf{Answer:} 3 wins, 7 losses

\textbf{Reasoning:}
\begin{enumerate}[label=\arabic*.]
    \item Question Independence: true - Self-contained question.
    \item Answer Specificity: true - Specific win-loss record.
    \item Question Unambiguity: true - Clear what is being asked.
    \item Question Answerability: true - The question is clearly stated.
    \item Avoid Open-ended Questions: true - Requires a specific record.
    \item Avoid Oversimplicity: true - Requires searching.
    \item Time Specificity: true - Contains specific year ``2025''.
\end{enumerate}
All conditions are true, therefore the answer is \texttt{\textbackslash boxed\{true\}}.

\noindent\hrulefill

\textbf{Example 7}

\textbf{Question:} What is Frances Tiafoe's current career tour-level finals record?

\textbf{Answer:} 3 wins, 7 losses

\textbf{Reasoning:}
\begin{enumerate}[label=\arabic*.]
    \item Question Independence: true - Self-contained question.
    \item Answer Specificity: true - Specific win-loss record.
    \item Question Unambiguity: true - Clear what is being asked.
    \item Question Answerability: true - The question is clearly stated.
    \item Avoid Open-ended Questions: true - Requires a specific record.
    \item Avoid Oversimplicity: true - Requires searching.
    \item Time Specificity: false - ``current'' is vague.
\end{enumerate}
There is a false condition, therefore the answer is \texttt{\textbackslash boxed\{false\}}.

\noindent\hrulefill

\textbf{IMPORTANT:} If you think any question is a multiple-choice question or a yes/no question, answer false.

\textbf{Yes/no example:}

\textbf{Question:} \zh{在2007年9月15日发表于《Biological Psychiatry》第62卷第6期的论文中，研究者在使用卡比多巴处理大鼠脑片后，3,4-亚甲二氧基甲基苯丙胺（MDMA）诱导的放电抑制和膜超极化现象是否消失？}

\textbf{Answer:} \zh{消失}

This is a yes/no question, therefore the answer is \texttt{\textbackslash boxed\{false\}}.

Now evaluate the following QA pair and provide \textbf{Reasoning}:
\begin{lstlisting}[basicstyle=\ttfamily\footnotesize,breaklines=true]
Question: {question}
Answer:   {answer}
\end{lstlisting}
\end{promptbox}
\subsection{Multi-hop QA Synthesis}

In addition to the single-hop QA extraction pipeline described in Section~\ref{data_gen}, we employ a separate \textbf{backward graph-based} multi-hop QA synthesis method. The core idea is to first build a knowledge graph grounded in real web evidence, then sample a connected subgraph, and finally generate a question \emph{backward} from the graph structure so that answering it naturally requires multi-hop reasoning along the graph's edges. The full pipeline is summarized in Table~\ref{tab:multihop_pipeline}.

\begin{table}[!ht]
    \centering
    \small
    \setlength{\tabcolsep}{4pt}
    \renewcommand{\arraystretch}{1.25}
    \begin{tabular}{@{} c l p{9.8cm} @{}}
        \toprule
        \textbf{Phase} & \textbf{Step} & \textbf{Operation} \\
        \midrule
        \multirow{5}{*}{\rotatebox[origin=c]{90}{\parbox{2.8cm}{\centering\textbf{1. Knowledge}\\\textbf{Graph ($\le N_{\max}$)}}}}
        & \textbf{Input} & Seed entity $e_0$; hyperparams $N_{\max}\!=\!8$, $K_{\text{feat}}\!=\!2$, $K_{\text{ent}}\!=\!2$ \\
        & \textbf{Search} & Query Serper API for entity $e_i$; LLM selects reliable sources; crawl pages \\
        & \textbf{Extract} & LLM extracts $K_{\text{feat}}$ factual features per entity from crawled evidence \\
        & \textbf{Discover} & LLM identifies $\le K_{\text{ent}}$ new concrete entities + directed relations; exclude generic concepts (e.g., ``Programs of Study'') and media sources (e.g., ``Wikipedia'') \\
        & \textbf{Expand} & Add new entities to graph $\mathcal{G}$; repeat \textbf{Search}$\to$\textbf{Discover} until $|V| \ge N_{\max}$ \\
        \midrule
        \multirow{3}{*}{\rotatebox[origin=c]{90}{\parbox{1.8cm}{\centering\textbf{2. Subgraph}\\\textbf{Sampling}}}}
        & \textbf{Init} & Pick random node $v_0 \in \mathcal{G}$ as BFS root \\
        & \textbf{Grow} & Greedily add candidate with most edges to subgraph + perturbation $\text{Uniform}(0, 0.5)$ \\
        & \textbf{Format} & Anonymize entity IDs; truncate to $\le$3 features per entity; output 6-node subgraph \\
        \midrule
        \multirow{4}{*}{\rotatebox[origin=c]{90}{\parbox{2.2cm}{\centering\textbf{3. Backward}\\\textbf{QA Gen}}}}
        & \textbf{Select} & Strong LLM picks a target entity as the answer \\
        & \textbf{Constrain} & Convert each edge $\to$ relationship constraint with vague references (e.g., ``a late antique writer'') \\
        & \textbf{Augment} & Optionally add minimal entity constraints from features (only if needed for uniqueness) \\
        & \textbf{Compose} & Integrate all constraints into a natural-language multi-hop question \\
        \midrule
        & \textbf{Output} & QA pair requiring 3--5 hop reasoning; information-minimal design prevents trivial solving \\
        \bottomrule
    \end{tabular}
    \caption{\textbf{Multi-hop QA synthesis pipeline.} Three phases: (1) iteratively build a knowledge graph from web evidence, (2) sample a connectivity-biased subgraph, (3) generate a backward question that requires traversing the subgraph to answer.}
    \label{tab:multihop_pipeline}
\end{table}

\subsection{Corpus Evolution}

As described in Section~\ref{data_gen}, our corpus grows iteratively: for each validated QA pair, we use the question as a search query to fetch related webpages from the open internet, which are then added to the local corpus. Starting from an initial corpus of $\sim$10M webpages (Wikipedia and curated seed sources), we expand the corpus by $\sim$22M additional real web pages over two iterations, reaching a total of $\sim$32M webpages. The expansion requires $\sim$220K Serper API calls (each returning $\sim$100 results), costing only $\sim$\$220---a one-time investment that is negligible compared to the \$59K--\$243K that would be required if online APIs were used during RL training (Table~\ref{tab:local_vs_online}). Table~\ref{tab:corpus_stats} summarizes the corpus statistics at each stage.

\begin{table}[!ht]
    \centering
    \small
    \setlength{\tabcolsep}{6pt}
    \renewcommand{\arraystretch}{1.1}
    \begin{tabular}{lrrr}
        \toprule
        \textbf{Stage} & \textbf{\#Webpages} & \textbf{Web-crawled} & \textbf{Serper Calls} \\
        \midrule
        Initial corpus (Wiki + Cache Webpages) & $\sim$10M & 0 & 0 \\
        After iteration 1 & $\sim$21M & $\sim$11M & $\sim$110K \\
        After iteration 2 & $\sim$32M & $\sim$22M & $\sim$110K \\
        \cmidrule{2-4}
        \textit{Total} & \textit{$\sim$32M} & \textit{$\sim$22M} & \textit{$\sim$220K (\$220)} \\
        \bottomrule
    \end{tabular}
    \caption{\textbf{Corpus expansion statistics.} The initial 10M pages come from Wikipedia and curated seeds (zero API cost). The iterative expansion adds $\sim$22M real web pages via $\sim$220K Serper calls (\$220), after which all RL training runs at zero marginal cost.}
    \label{tab:corpus_stats}
\end{table}

\begin{figure}[!ht]
    \centering
    \includegraphics[width=\textwidth]{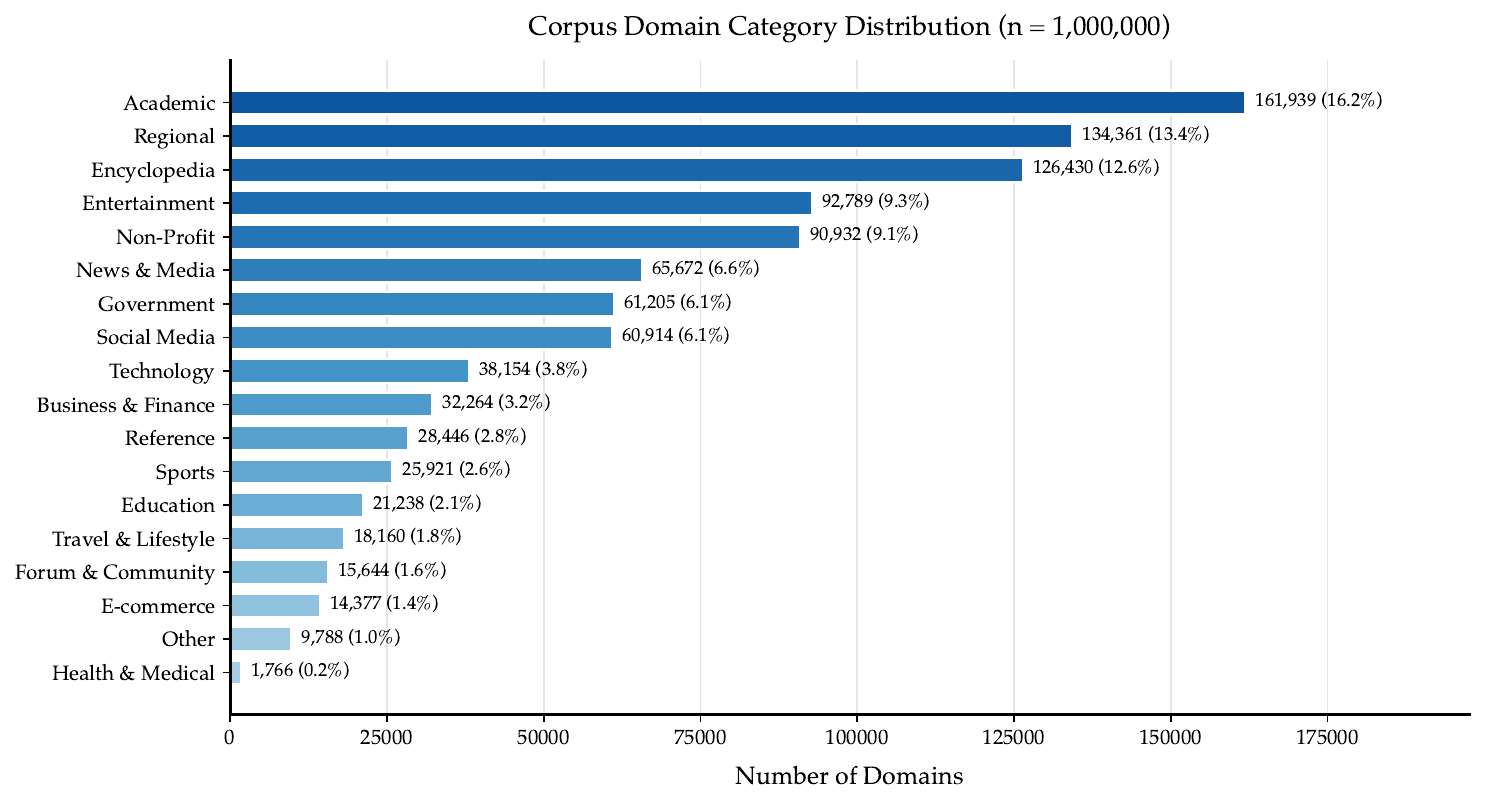}
    \caption{\textbf{Corpus domain category distribution.} The enriched corpus spans 18 domain categories covering 1M+ unique domains, with Academic, Regional, and Encyclopedia sources forming the largest segments. This broad coverage ensures that the local search environment reflects diverse real-world web structure.}
    \label{fig:domain_dist}
\end{figure}

Each fetched webpage undergoes deduplication (URL-level and content-hash), HTML-to-Markdown conversion, and minimum length filtering ($>$1000 characters). The crawled pages are simultaneously indexed into the local search engine (Milvus) and stored in PostgreSQL for the local browse tool, ensuring that every new page is immediately available for both RL rollouts and subsequent QA synthesis iterations. Figure~\ref{fig:domain_dist} shows the domain category distribution of the final enriched corpus.

\section{SFT Details}
\label{sec:app_sft}

\subsection{Data Composition}

The SFT dataset consists of 68,231 high-quality search trajectories from three categories (Table~\ref{tab:sft_data_sources}): our co-constructed synthesis pipeline contributes the majority (52.8\%), supplemented by multi-hop QA data (7.1\%) and trajectories distilled from open-source QA pairs (40.2\%). All trajectories are generated by running Tongyi DeepResearch \citep{team2025tongyi} as the teacher model on each QA pair with online search tools; we use only the QA pairs from open-source datasets, not their original trajectories.

\begin{table}[!ht]
    \centering
    \small
    \setlength{\tabcolsep}{6pt}
    \renewcommand{\arraystretch}{1.1}
    \begin{tabular}{llrr}
        \toprule
        \textbf{Category} & \textbf{Source} & \textbf{\#Samples} & \textbf{Proportion} \\
        \midrule
        \multirow{1}{*}{Our synthesis (Sec.~\ref{data_gen})} & Direct information queries & 36,009 & 52.8\% \\
        \midrule
        \multirow{2}{*}{Multi-hop QA (App.~A.3)} & Syn from English seed & 2,311 & 3.4\% \\
                                                   & Syn from Chinese seed & 2,502 & 3.7\% \\
        \midrule
        \multirow{3}{*}{Open-source QA distillation} & MiroRL-GenQA~\citep{2025mirorl} & 10,400 & 15.2\% \\
                                                   & ASearcher~\citep{gao2025beyond} & 8,364 & 12.3\% \\
                                                   & TaskCraft~\citep{shi2025taskcraft} & 8,645 & 12.7\% \\
        \midrule
        \multicolumn{2}{l}{\textbf{Total}} & \textbf{68,231} & \textbf{100\%} \\
        \bottomrule
    \end{tabular}
    \caption{\textbf{SFT data composition.}}
    \label{tab:sft_data_sources}
\end{table}

\subsection{Trajectory Processing}

Raw trajectories from the teacher model undergo a two-stage processing pipeline (Table~\ref{tab:sft_pipeline}): seven filtering rules remove low-quality trajectories, and four cleaning steps standardize the surviving data.

\begin{table}[!ht]
    \centering
    \footnotesize
    \setlength{\tabcolsep}{4pt}
    \renewcommand{\arraystretch}{1.15}
    \begin{tabular}{@{} l l p{8.5cm} @{}}
        \toprule
        \textbf{Stage} & \textbf{Step} & \textbf{Criterion / Operation} \\
        \midrule
        \multirow{7}{*}{\textbf{Filter}}
        & F1 & Keep only trajectories with correct final answers (LLM judge) \\
        & F2 & Remove if same tool call repeated $>$3 times \\
        & F3 & Remove if fewer than 2 tool calls \\
        & F4 & Remove if contains \texttt{PythonInterpreter} \\
        & F5 & Remove if contains encoding artifacts or \texttt{\textbackslash boxed\{\}} \\
        & F6 & Remove if $>$2 tool-call errors (timeouts, access failures) \\
        & F7 & Remove if tool call issued without preceding \texttt{<think>} tag \\
        \midrule
        \multirow{4}{*}{\textbf{Clean}}
        & C1 & Replace system prompt with standardized format \\
        & C2 & Shorten final answer to reference if $>$100 characters \\
        & C3 & Merge intermediate reminder prompts into adjacent turn \\
        & C4 & Truncate to end immediately after \texttt{</answer>} tag \\
        \bottomrule
    \end{tabular}
    \caption{\textbf{SFT trajectory processing pipeline.}}
    \label{tab:sft_pipeline}
\end{table}

\begin{figure}[!ht]
    \centering
    \includegraphics[width=\textwidth]{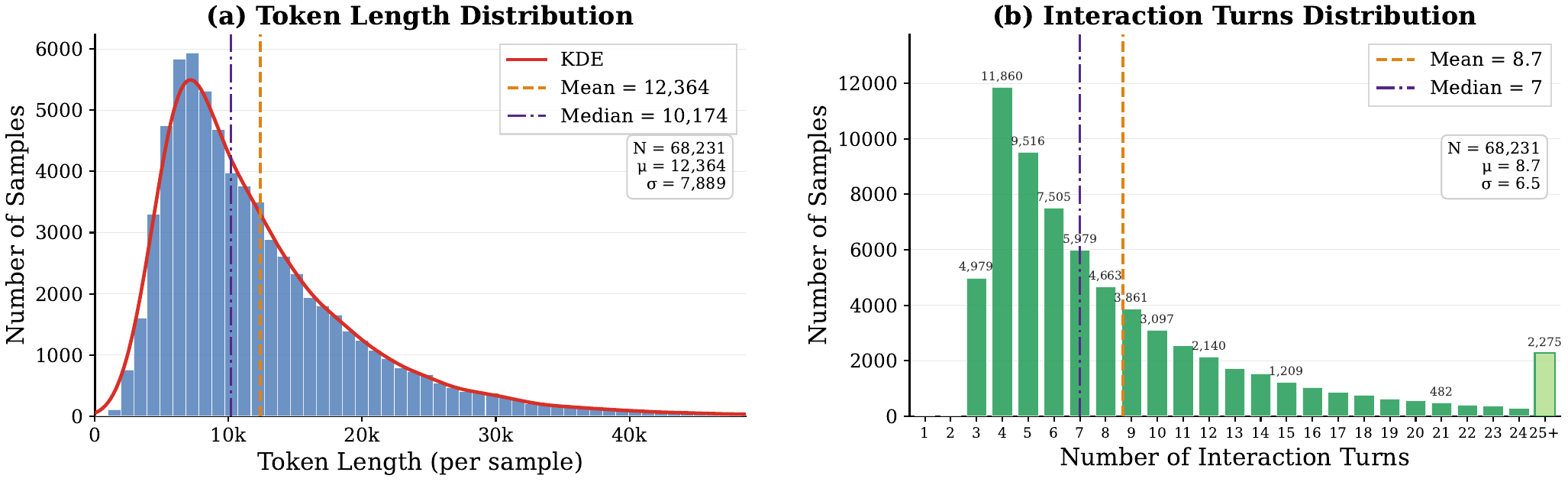}
    \caption{\textbf{SFT data distribution (after processing).} \textbf{(a)} Token length (mean 12.4K, median 10.2K). \textbf{(b)} Interaction turns (mean 8.7, median 7).}
    \label{fig:sft_data_dist}
\end{figure}

Figure~\ref{fig:sft_data_dist} shows the distribution of the final 68K trajectories after processing: the mean token length is 12.4K with a long tail extending to $\sim$45K, and the mean number of interaction turns is 8.7, concentrated around 5--8 turns. The long tail motivates our choice of 64K max sequence length to cover 100\% of samples.

\subsection{Training and Baseline Performance}

We initialize from Qwen3-4B-Thinking-2507 and perform full-parameter SFT using LLaMA-Factory. All trajectories are formatted using the Qwen3 thinking template with \texttt{enable\_thinking=true}. Training uses a maximum sequence length of 64K tokens, a learning rate of $2 \times 10^{-5}$ with cosine scheduling and 10\% warmup, and runs for 1 epoch on 8$\times$H100 GPUs with DeepSpeed ZeRO-2, Flash Attention 2, and Liger Kernel optimizations. The effective batch size is $2 \times 8 \times 8 = 128$.

Table~\ref{tab:sft_baseline} shows the performance progression. SFT provides a strong initialization for RL, but leaves substantial room for improvement---particularly on benchmarks requiring complex multi-step reasoning.

\begin{table}[!ht]
    \centering
    \footnotesize
    \setlength{\tabcolsep}{2pt}
    \renewcommand{\arraystretch}{1.1}
    \begin{tabular}{c cccc cccc}
        \toprule
        \textbf{Model} & 
        \begin{tabular}{@{}c@{}}\textbf{GAIA-} \\ \textbf{Text}\end{tabular} & 
        \begin{tabular}{@{}c@{}}\textbf{Browse-} \\ \textbf{comp}\end{tabular} & 
        \begin{tabular}{@{}c@{}}\textbf{Browse.} \\ \textbf{(ZH)}\end{tabular} & 
        \textbf{HLE} & 
        \textbf{Frames} & 
        \begin{tabular}{@{}c@{}}\textbf{Web-} \\ \textbf{walker}\end{tabular} & 
        \textbf{Seal-0} & 
        \begin{tabular}{@{}c@{}}\textbf{Xbench-} \\ \textbf{DS}\end{tabular} \\
        \midrule
        Base Model & 28.16\% & 1.8\% & 8.3\% & 9.12\% & 36.77\% & 29.6\% & 7.9\% & 21.0\% \\
        After SFT & 55.58\% & 9.5\% & 14.88\% & 17.23\% & 71.24\% & 60.88\% & 30.18\% & 64.25\% \\
        \midrule
        \rowcolor{blue!8} After RL & \textbf{71.3\%}\textsuperscript{\scriptsize\textcolor{red!70!black}{$\uparrow$15.7}} & \textbf{20.3\%}\textsuperscript{\scriptsize\textcolor{red!70!black}{$\uparrow$10.8}} & \textbf{28.4\%}\textsuperscript{\scriptsize\textcolor{red!70!black}{$\uparrow$13.5}} & \textbf{22.0\%}\textsuperscript{\scriptsize\textcolor{red!70!black}{$\uparrow$4.8}} & \textbf{83.1\%}\textsuperscript{\scriptsize\textcolor{red!70!black}{$\uparrow$11.9}} & \textbf{72.7\%}\textsuperscript{\scriptsize\textcolor{red!70!black}{$\uparrow$11.8}} & \textbf{41.8\%}\textsuperscript{\scriptsize\textcolor{red!70!black}{$\uparrow$11.6}} & \textbf{78.0\%}\textsuperscript{\scriptsize\textcolor{red!70!black}{$\uparrow$13.8}} \\
        \bottomrule
    \end{tabular}
    \caption{\textbf{Performance progression from base model through SFT to RL. All experiments use a 128k context length (without memory).}}
    \label{tab:sft_baseline}
\end{table}

\section{RL Details}
\label{sec:app_rl}

\subsection{Data Mixture}

Table~\ref{tab:rl_data_mixture} summarizes the data composition used in each RL stage. Stage~1 uses a simple two-source mixture dominated by direct QA, while Stage~2 diversifies the distribution with wiki-sourced queries, multi-hop subgraph reasoning, and science-domain data to push beyond the Stage~1 plateau. All queries undergo difficulty-aware filtering (pass@8, $1 \le c \le 7$) before being fed to training.

\begin{table}[!ht]
    \centering
    \small
    \setlength{\tabcolsep}{6pt}
    \renewcommand{\arraystretch}{1.1}
    \begin{tabular}{lrr}
        \toprule
        \textbf{Data Source} & \textbf{\#Queries} & \textbf{Proportion} \\
        \midrule
        \multicolumn{3}{l}{\textit{\textbf{Stage 1}}} \\
        Our synthetic data & 7,634 & 73.4\% \\
        Multi-hop QA & 2,764 & 26.6\% \\
        \cmidrule{2-3}
        \textit{Total (Stage 1)} & \textit{10,398} & \textit{100\%} \\
        \midrule
        \multicolumn{3}{l}{\textit{\textbf{Stage 2}}} \\
        Our synthetic data & 11,101 & 68.6\% \\
        Multi-hop QA & 3,298 & 20.3\% \\
        Science~\citep{fan2025megascience} & 1,800 & 11.1\% \\
        \cmidrule{2-3}
        \textit{Total (Stage 2)} & \textit{16,199} & \textit{100\%} \\
        \bottomrule
    \end{tabular}
    \caption{\textbf{RL training data mixture per stage} (after difficulty-aware filtering). Stage~2 introduces broader domain coverage and higher complexity.}
    \label{tab:rl_data_mixture}
\end{table}

\subsection{Reward Function}
\label{sec:reward_prompt}

During RL training, each rollout receives a binary reward based on the correctness of the agent's final answer. The agent is required to produce its answer within \texttt{<answer>...</answer>} tags. We use \texttt{Qwen3-30B-A3B-Instruct} as the reward judge for LLM-based semantic judgment.

\begin{promptbox}{LLM Judge Evaluation Prompt}
You are an evaluation assistant. Please determine if the predicted answer is semantically equivalent to the labeled answer.\\[1ex]
\textbf{Question:} \texttt{\{question\}}\\[0.5ex]
\textbf{Labeled Answer:} \texttt{\{correct\_answer\}}\\[0.5ex]
\textbf{Predicted Answer:} \texttt{\{response\}}\\[1ex]
Please evaluate the answer and return a JSON object with the following format:\\[0.5ex]
\verb|{|\verb|"reasoning": "...", "judgment": "Correct"|\verb|}|\\[0.5ex]
If the answers are not equivalent, the ``judgment'' field should be ``Incorrect''.\\[0.5ex]
Output ONLY the JSON object, without any markdown formatting or additional text.
\end{promptbox}

\subsection{RL Training Configuration}
\label{sec:app_rl_config}

Table~\ref{tab:rl_config} summarizes the full hyperparameter configuration for our RL training pipeline. We use SGLang as the rollout (inference) engine and FSDP as the training engine.

\begin{table}[!ht]
    \centering
    \footnotesize
    \setlength{\tabcolsep}{4pt}
    \renewcommand{\arraystretch}{1.15}
    \begin{tabularx}{\textwidth}{@{} l l X @{}}
        \toprule
        \textbf{Category} & \textbf{Hyperparameter} & \textbf{Value} \\
        \midrule
        \multirow{7}{*}{\textbf{Optimization}}
        & Algorithm & GRPO (on-policy, single update per rollout batch) \\
        & Global batch size & 128 queries \\
        & Rollouts per query ($K$) & 8 \\
        & Mini-batch size & 128 (= batch size, strictly on-policy) \\
        & Learning rate & $1 \times 10^{-6}$ (constant) \\
        & KL penalty / Entropy bonus & None / None \\
        & Loss aggregation & token-mean, sequence-mean \\
        \midrule
        \multirow{4}{*}{\textbf{Rollout}}
        & Max prompt length & 1,024 tokens \\
        & Max response length & 32K (Stage~1) $\to$ 48K (Stage~2) \\
        & Sampling temperature & 0.7 (Stage~1) / 1.0 (Stage~2) \\
        & Top-$p$ & 0.95 \\
        \midrule
        \multirow{1}{*}{\textbf{Multi-turn Agent}}
        & Max assistant turns & 40 (Stage~1) / 60 (Stage~2) \\
        \midrule
        \multirow{3}{*}{\textbf{TIS Correction}}
        & IS threshold (upper / lower) & 2.0 / 0.5 (auto) \\
        & Aggregation level / mode & Token-level / Truncate \\
        & Veto threshold & $1 \times 10^{-4}$ \\
        \bottomrule
    \end{tabularx}
    \caption{\textbf{RL training configuration.} All RL experiments are conducted using the VERL framework with Ray-based distributed orchestration. TIS (Trajectory Importance Sampling) corrects the distribution mismatch between the SGLang rollout engine (BF16) and the FSDP training engine (FP32).}
    \label{tab:rl_config}
\end{table}

\subsection{Trajectory Behavior During RL}

After SFT, the model acquires the basic search workflow and achieves a clear performance gain over the base model. However, we observe that among the incorrectly answered queries, a dominant failure mode is \emph{repetitive action loops}---the model repeatedly issues the same search query or revisits the same URL without making progress, consuming the token budget on redundant interactions rather than exploring alternative strategies. These repetitive trajectories inflate both the response length and the number of turns without contributing useful information.

\begin{figure}[!ht]
    \centering
    \includegraphics[width=\textwidth]{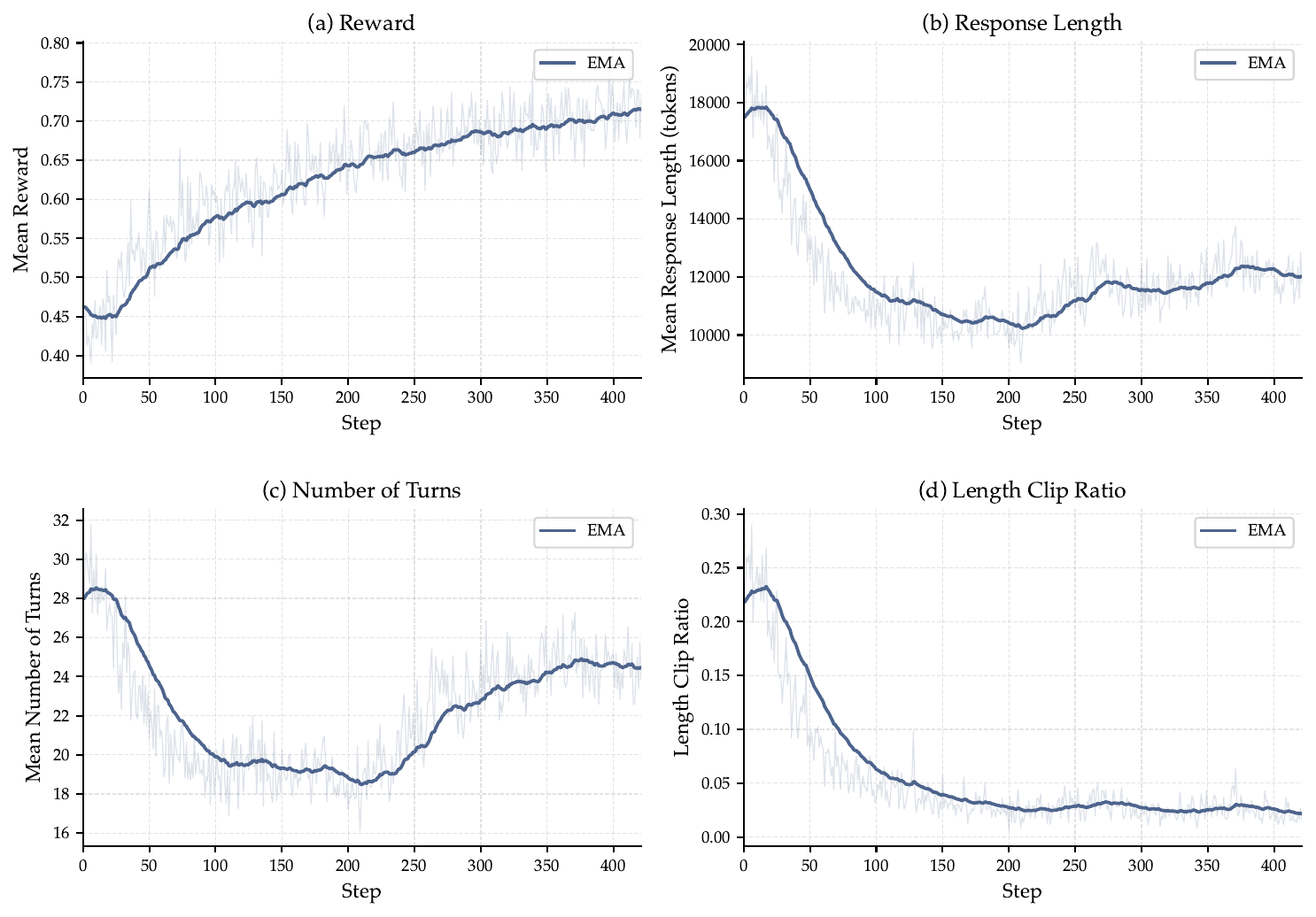}
    \caption{\textbf{RL suppresses repetitive actions inherited from SFT.} \textbf{(a)} Mean reward increases from $\sim$0.42 to $\sim$0.70, confirming improved task accuracy. \textbf{(b--d)} Mean response length ($\sim$18K$\to$12K tokens), interaction turns ($\sim$30$\to$24), and length clip ratio ($\sim$0.28$\to$0.02) all decrease, reflecting elimination of redundant action loops. No explicit length or repetition penalty is used.}
    \label{fig:rl_dynamics}
\end{figure}

Figure~\ref{fig:rl_dynamics} tracks four metrics during RL training that jointly show how the model eliminates this failure mode while improving task performance.

The mean reward steadily increases from $\sim$0.42 to $\sim$0.70 over 420 steps (Figure~\ref{fig:rl_dynamics}a), confirming that the agent solves progressively more queries correctly. The remaining three metrics reveal the mechanism behind this improvement: as RL suppresses the repetitive-action loops, the mean response length drops from $\sim$18K to $\sim$12K tokens (Figure~\ref{fig:rl_dynamics}b), the mean number of interaction turns decreases from $\sim$30 to $\sim$24 (Figure~\ref{fig:rl_dynamics}c), and the length clip ratio---the fraction of trajectories truncated for exceeding the context window---falls from $\sim$0.28 to $\sim$0.02 (Figure~\ref{fig:rl_dynamics}d). All three reductions reflect the same underlying cause: the model learns to stop repeating futile actions, resulting in shorter, more purposeful search trajectories.

Notably, this behavioral correction emerges purely from the outcome-based reward signal and the GRPO clipping mechanism, without any explicit length penalty or repetition penalty in the loss function.

\subsection{Training Dynamics}

We track several metrics across the full two-stage RL training process. Stage~1 (steps 0--220, temperature 0.7) trains on local-RAG-only data, and Stage~2 (from step~220, temperature 1.0) continues from the Stage~1 checkpoint with the full data mixture.

\begin{figure}[!ht]
    \centering
    \includegraphics[width=\textwidth]{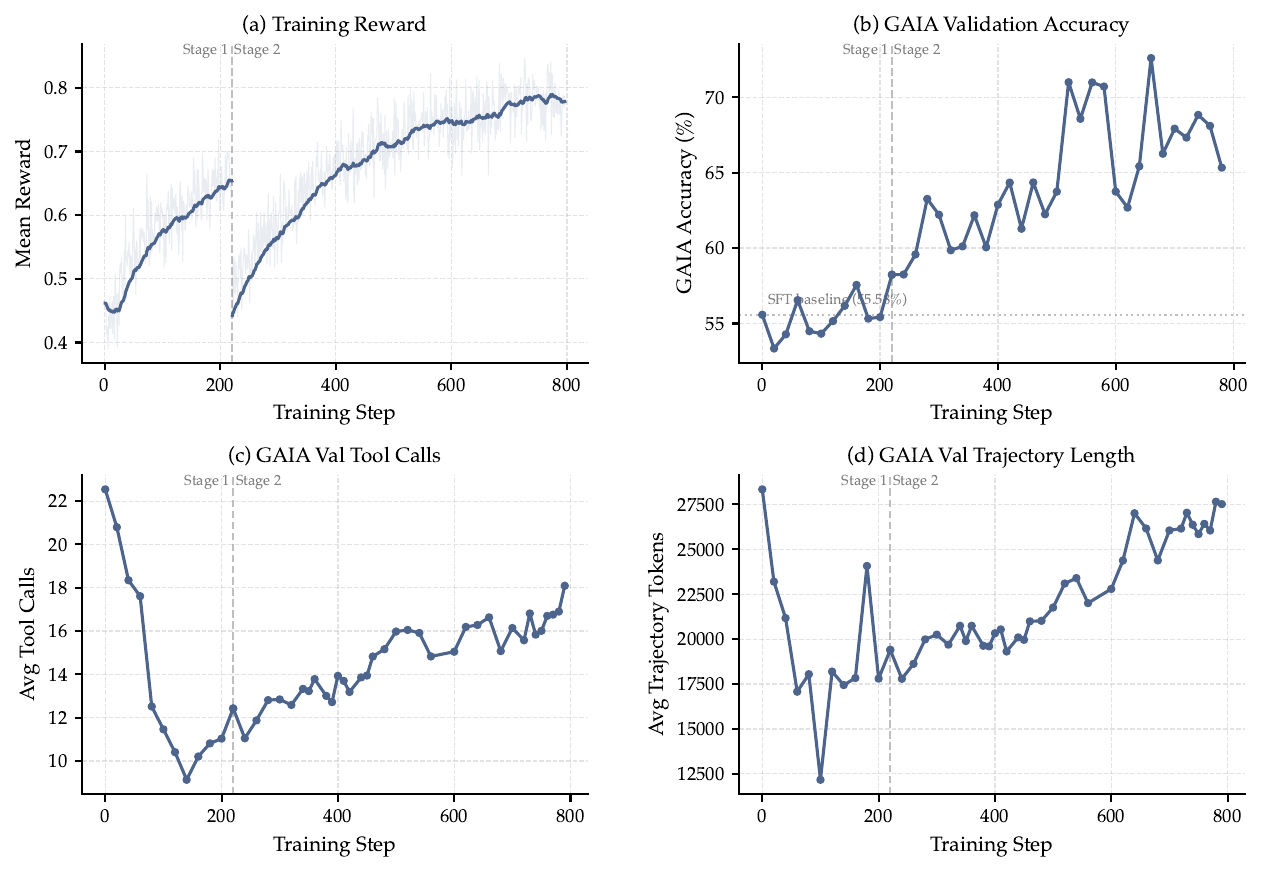}
    \caption{\textbf{Training dynamics during RL.} (a)~GAIA validation accuracy. (b)~Policy entropy (Stage~1: temp\,=\,0.7; Stage~2: temp\,=\,1.0). (c)~Average tool calls per sample. (d)~Average trajectory total tokens. Dashed vertical lines mark the Stage~1$\to$2 transition at step~220.}
    \label{fig:training_dynamics}
\end{figure}

\paragraph{Training dynamics.}
Figure~\ref{fig:training_dynamics} summarizes four key metrics across the full two-stage RL process.
Panel~(a) shows the GAIA validation accuracy: Stage~1 establishes the initial performance gain over the SFT baseline, and Stage~2 further improves accuracy, reaching a peak of $\sim$68\%.
Panel~(b) shows the policy entropy; Stage~2 uses a higher sampling temperature, resulting in higher absolute entropy, while entropy decreases steadily in both stages, indicating increasingly confident action selection.
Panels~(c) and~(d) jointly plot the average tool calls and trajectory total tokens. During Stage~1, both metrics drop sharply as the model eliminates redundant action loops. In Stage~2, both increase together, indicating the model performs more search steps and generates longer but more productive trajectories.

\section{Infrastructure Details}
\label{sec:app_infra}

Table~\ref{tab:infra_config} summarizes the key configuration of our local search and browse services.

\begin{table}[!ht]
    \centering
    \footnotesize
    \setlength{\tabcolsep}{3pt}
    \renewcommand{\arraystretch}{1.15}

    \begin{tabularx}{\textwidth}{@{} l l X @{}}
        \toprule
        \textbf{Service} & \textbf{Parameter} & \textbf{Value / Description} \\
        \midrule
        \multirow{7}{*}{\textbf{Search}}
        & Version & Milvus v2.6.0 standalone (MinIO + etcd) \\
        & Embedding & BGE-M3: 1024-d dense + learned sparse per page \\
        & Index type & DiskANN with mmap \\
        & MaxDegree & 64 \\
        & SearchListSize & 128 \\
        & SearchCache\-BudgetGBRatio & 0.9 ($\sim$200\,GB host memory as search cache) \\
        & Hybrid fusion & Reciprocal Rank Fusion (RRF) at query time \\
        \midrule
        \multirow{6}{*}{\textbf{Browse}}
        & Content format & Full Markdown, keyed by URL \\
        & max\_connections & 1000 \\
        & shared\_buffers & 1\,GB \\
        & effective\_cache\_size & 4\,GB \\
        & work\_mem & 4\,MB (bounded for 1K concurrent queries) \\
        & Storage & NFS-backed NAS, B-tree index on URL \\
        \bottomrule
    \end{tabularx}
    \caption{\textbf{Infrastructure configuration} for the local search engine and browse tool.}
    \label{tab:infra_config}
\end{table}

\section{Example Trajectories}
\label{sec:app_traj}

To illustrate the qualitative behavior of \textbf{LiteResearcher-4B}, we present four additional representative rollouts (the BrowseComp case is in the main text, Section~\ref{sec:analysis}), all automatically judged correct and manually verified to contain no answer leakage. Each raw trajectory spans tens to over a hundred interaction steps; we render them \emph{turn by turn}, quoting the agent's own reasoning (\textcolor{trajthink}{\sffamily\scriptsize\bfseries thought}) alongside its actions (\textcolor{trajsearch}{\sffamily\scriptsize\bfseries SEARCH}, \textcolor{trajvisit}{\sffamily\scriptsize\bfseries VISIT}) and the resulting observations (\textcolor{simplexmuted}{\sffamily\scriptsize\bfseries obs}), abbreviating runs of similar steps (e.g.\ ``TURN 8--13'').

\begin{table}[h]
\centering
\footnotesize
\setlength{\tabcolsep}{5pt}
\renewcommand{\arraystretch}{1.25}
\begin{tabularx}{\textwidth}{@{} p{0.27\textwidth} l X @{}}
\toprule
\textbf{Clue} & \textbf{Source found by agent} & \textbf{Evidence} \\
\midrule
(i)~University instructor & Wikipedia & ``former university English teacher'' \\
(ii)~Translated into 20+ languages & author / publisher pages & translated into \textbf{28} languages (e.g.\ \emph{Wicked Lovely}) \\
(iii)~Worked as an editor & interviews & edited anthologies (co-edited with Kelley Armstrong) \\
(iv)~Parents in the NRA & Lightspeed Magazine & ``a father who collected guns and was an NRA member'' \\
(v)~Wolf nightmare & Goodreads interview & ``a recurring nightmare \ldots about a wolf'' \\
(vi)~Three children, two adopted & Publishers Weekly, interviews & three children, two of them adopted \\
\bottomrule
\end{tabularx}
\caption{\textbf{BrowseComp case study (main text, Section~\ref{sec:analysis}): clue\,$\to$\,evidence verification.} The agent anchored each of the six scattered constraints to an \emph{independent} source; because no single page contained the answer, the identification required genuine multi-hop synthesis rather than a lookup.}
\label{tab:browsecomp_clues}
\end{table}

\paragraph{Xbench --- cross-domain multi-hop with manual date arithmetic.}
This Chinese-language query chains four domains --- a quote, the novel it appears in, that novel's film adaptation, and an Olympic record --- and ends in a date subtraction. With no code interpreter available, the agent tries several online date calculators (all of which need interactive input and return nothing), then computes the difference \emph{by hand} with leap-year rules and cross-checks its own result.

\begin{trajbox}{Xbench: quote $\to$ novel $\to$ film $\to$ Olympic record $\to$ day count (judged correct)}
\textbf{Question.}\ \zh{语句"如山间清爽的风 如古城温暖的光"所出自的小说，被拍成电影，其中电影上映时间距离中国的第一枚奥运会赛艇金牌已经过去了多少天？}

{\small\itshape English: how many days passed between China's first Olympic rowing gold and the release of the film adapted from the novel containing the line ``as a cool mountain breeze, as the warm light of an ancient town''? (Think excerpts translated from the agent's original Chinese.)}

\smallskip
\textbf{Reference:}\ \zh{2965天} (2965 days) \hfill \textbf{Model answer:}\ \textbf{2965}~\textcolor{trajsearch}{\faCheck}
\smallskip

\trturn{1}
\trquote{I must find which novel the line is from, then its film adaptation and that film's release date; then the date of China's first Olympic rowing gold; then subtract the two dates.}
\trsearch{\zh{如山间清爽的风 如古城温暖的光} \textnormal{(the quoted line)}}
\trobs{\textnormal{The line is from Zhang Jiajia's novel} \zh{《从你的全世界路过》}.}

\trturn{2--4}
\trquote{The novel was adapted into a same-name film. The English Wikipedia title won't load --- I'll switch to the Chinese Wikipedia.}
\trvisit{\texttt{zh.wikipedia.org/wiki/}\zh{从你的全世界路过}}
\trobs{\textnormal{Release dates:} \textbf{2016-09-29} \textnormal{(Mainland China), 09-30 (US), 10-07 (UK). I take the Mainland date.}~\textcolor{trajsearch}{\faCheck}}

\trturn{5--6}
\trquote{Now I need the date of China's first Olympic rowing gold medal.}
\trsearch{\zh{中国第一枚奥运会赛艇金牌}}
\trvisit{\texttt{sport.gov.cn}}
\trobs{\textnormal{Women's quadruple sculls at Beijing 2008; the article is dated Aug 18 but reports the race on Beijing time Aug 17.}}

\trturn{7}
\trquote{The page is dated 2008-08-18 but says the race was on Beijing time Aug 17. So the medal date is \textbf{2008-08-17}, and I must count the days from 2008-08-17 to 2016-09-29.}

\trturn{8}
\trquote{2008 is divisible by 4 and not a century year, so it is a leap year. I'll segment the interval year by year starting from Aug 17, with the leap years 2008/2012/2016 contributing 366 days each.}

\trturn{11}
\trquote{Year segments from 2008-08-17 to 2016-08-17: $366 + 365 + 365 + 365 + 366 + 365 + 365 + 365$. Running sum: $731, 1096, 1461, 1827, 2192, 2557, \mathbf{2922}$. So 2922 days to 2016-08-17.}
\trsearch{\zh{2008年8月17日 到 2016年9月29日 天数}}
\trobs{\textnormal{Only links to interactive date calculators.}}

\trturn{13--14}
\trvisit{\texttt{calculator-online.net/...\,days-since}}
\trquote{This calculator's summary says \textbf{2983} days. Is that right? Let me cross-check against my own decomposition.}

\trturn{15}
\trquote{2008-08-17 to 2016-08-17 is \emph{exactly} 8 years containing two leap days (2012, 2016), so $8\times 365 + 2 = 2922$; then 2016-08-17 to 2016-09-29 adds $14 + 30 + 29 = 43$ days, giving \textbf{2965}. The calculator's 2983 is inconsistent with the explicit leap-year count, so 2983 must be wrong.}

\trturn{16--18}
\trvisit{\texttt{timeanddate.com}, \texttt{gigacalculator.com} \textnormal{--- even injecting} \texttt{?start\_date=...\&end\_date=...}}
\trobs{\textnormal{All require interactive form input; none returns a usable number, so the hand computation stands.}}

\transwer{$8\times 365 + 2 + 43 = 2922 + 43 = \mathbf{2965}$ \textnormal{days; the self-check discards the erroneous 2983.}}
\end{trajbox}

\paragraph{GAIA --- multi-source verification with exact arithmetic.}
The agent resolves a $1002$ vs.\ $1037$ source discrepancy, discards an unreliable page, then computes the answer exactly.

\begin{trajbox}{GAIA: multi-source verification with exact arithmetic (judged correct)}
\textbf{Question.}\ If we assume all articles published by \emph{Nature} in 2020 (articles only, not book reviews/columns) relied on statistical significance and on average reached a $p$-value of $0.04$, how many papers would be incorrect in their claims of statistical significance? Round up to the next integer.

\smallskip
\textbf{Reference:}\ 41 \hfill \textbf{Model answer:}\ \textbf{41}~\textcolor{trajsearch}{\faCheck}
\smallskip

\trturn{1}
\trquote{The question asks how many \emph{Nature}-2020 articles would be ``incorrect as to their claims of statistical significance''. I need two things: the number of articles, and a false-positive rate implied by the average $p$-value of 0.04.}
\trsearch{\texttt{Number of articles published by Nature in 2020} \textnormal{(+ 3 paraphrases)}}
\trobs{\textnormal{Nature's listing page shows} ``\texttt{Article (1002)}''.}

\trturn{2}
\trvisit{\texttt{nature.com/nature/articles?year=2020} \textnormal{--- goal: the 2020 article count}}
\trobs{\textnormal{Confirms the} ``\texttt{Article (1002)}'' \textnormal{filter entry.}}

\trturn{3}
\trquote{The page shows ``Article (1002)''. But the question says ``articles only, not book reviews/columns'', so I must be careful which category 1002 actually counts.}
\trvisit{\texttt{en.wikipedia.org/wiki/Nature\_(journal)}}
\trobs{\textnormal{Wikipedia gives an acceptance rate but} \emph{not} \textnormal{the 2020 count.}}

\trturn{4}
\trvisit{\texttt{nature.com/nature/research-articles}}
\trobs{\textnormal{This listing instead shows} \texttt{2020 (1037)} \textnormal{--- a} \emph{different} \textnormal{number.}}

\trturn{5}
\trquote{Now I have two different numbers, 1002 vs.\ 1037. Which is correct? ``Articles, only, not book reviews/columns'' points to the \texttt{Article}-type filter, which is 1002.}
\trsearch{\texttt{Nature 2020 article count 1002}}
\trobs{\textnormal{Returns scattered article DOIs, no clean count --- inconclusive.}}

\trturn{6}
\trvisit{\texttt{nature.com/...?type=article\&year=2020}}
\trobs{\textnormal{The} \texttt{type=article} \textnormal{filter explicitly reads} \texttt{Article (1002)}, \texttt{2020 (1002)}.~\textcolor{trajsearch}{\faCheck}}

\trturn{7}
\trquote{So the article-type count is 1002. Now the false-positive part: with an average $p$-value of 0.04, the natural reading is the product $1002 \times 0.04$.}
\trsearch{\texttt{p-value 0.04 false positive rate}}
\trobs{\textnormal{Statistics explainers, but no plug-in number.}}

\trturn{8}
\trvisit{\texttt{graphpad.com/...\,false-positive} \textnormal{FAQ}}
\trobs{\textnormal{Discusses false-positive risk but gives no closed form for $p\,{=}\,0.04$.}}

\trturn{9}
\trquote{The question expects the simple product $1002 \times 0.04 = 40.08$. ``Round up to the next integer'' is the ceiling, so $40.08 \Rightarrow 41$.}

\trturn{10--11}
\trquote{Could the intended count be 1037 (``research articles'')? But ``articles only'' matches the 1002 filter. One third-party page claims ``12,000+'' --- that counts \emph{all} content types, so it is unreliable and I discard it.}
\trvisit{\textnormal{third-party stats page --- claims ``12,000+'' (rejected)}}
\trobs{\textnormal{Unofficial; conflates all content categories --- discarded.}}

\trturn{12}
\trthink{Settles on 1002 article-type entries and applies the ceiling.}
\transwer{$\lceil 1002 \times 0.04 \rceil = \lceil 40.08 \rceil = \mathbf{41}$}
\end{trajbox}

\paragraph{Seal-0 --- exhaustive enumeration under a precise condition.}
Seal-0 rewards answers that survive an exhaustive check. Here the agent must count the former U.S.\ presidents satisfying a precise compound condition --- lost a general election, ran \emph{again}, and lost the second attempt --- which turns on edge cases (\emph{seeking a nomination} vs.\ \emph{being a candidate}). The agent enumerates every plausible president, verifies each against primary sources, and rules out the near-misses.

\begin{trajbox}{Seal-0: counting presidents under a precise compound condition (judged correct)}
\textbf{Question.}\ How many former US presidents ran again for president after losing a general election and lost on their second attempt?

\smallskip
\textbf{Reference:}\ 2 \hfill \textbf{Model answer:}\ \textbf{2} --- Van Buren, Fillmore~\textcolor{trajsearch}{\faCheck}
\smallskip

\trturn{1}
\trquote{I need former presidents who (a) lost a general election, (b) ran \emph{again}, and (c) lost that second attempt. Let me enumerate the candidates and check each against all three conditions.}
\trsearch{\texttt{former US president ran again after losing a general election, lost second attempt}}
\trobs{\textnormal{Wikipedia lists of presidents who lost reelection / ran for office again.}}

\trturn{4}
\trvisit{\texttt{en.wikipedia.org/wiki/Martin\_Van\_Buren}}
\trquote{Van Buren lost the 1840 general election, then ran again in 1848 as the Free Soil nominee and lost. All three conditions hold --- he counts.}

\trturn{5}
\trvisit{\texttt{en.wikipedia.org/wiki/Millard\_Fillmore}}
\trquote{Fillmore ran in 1856 as the American (Know-Nothing) nominee and lost the general election --- he counts too.}

\trturn{6 (rule out)}
\trvisit{\texttt{en.wikipedia.org/wiki/Herbert\_Hoover}}
\trquote{Hoover lost in 1932 and \emph{sought} the 1940 nomination --- but he was never a general-election candidate again. ``Ran for president'' should mean \emph{being a candidate}, so I exclude him.}

\trturn{12 (rule out)}
\trvisit{\texttt{theconversation.com/...\,former-presidents-running}}
\trquote{Theodore Roosevelt ran in 1912 and lost, but he had \emph{not} lost a general election beforehand --- he fails the ``after losing a general election'' clause.}

\trturn{15--17 (rule out)}
\trvisit{\texttt{en.wikipedia.org/wiki/Ulysses\_S.\_Grant}; \texttt{.../William\_Henry\_Harrison}}
\trquote{Grant only failed a \emph{nomination} in 1880, never a general-election candidate again; W.H.\ Harrison died in office. Neither qualifies.}

\trturn{19--22 (cross-verify)}
\trvisit{\texttt{pewresearch.org/...}; \texttt{en.wikipedia.org/wiki/1856\_United\_States\_presidential\_election}}
\trquote{Pew and the 1840/1856 election pages independently confirm Van Buren and Fillmore as the only two former presidents who lost a \emph{second} general-election bid.~\textcolor{trajsearch}{\faCheck}}

\trturn{35}
\trthink{Only Van Buren and Fillmore satisfy all three conditions; the near-misses (Hoover, T.\ Roosevelt, Grant, W.H.\ Harrison) each fail the precise wording.}
\transwer{\textbf{2} --- Martin Van Buren and Millard Fillmore.}
\end{trajbox}

\paragraph{HLE --- falling back to a first-principles derivation.}
Humanity's Last Exam (HLE) poses problems with no ready-made answer online. This one asks for an \emph{exact} side length in a geometry construction. After eight searches surface only loosely related discussions, the agent stops searching and \emph{derives the answer itself}: it adopts the method from a related two-square problem, sets up coordinates, and solves for $c$ in closed form.

\begin{trajbox}{HLE: an exact-answer geometry problem solved by self-derivation (judged correct)}
\textbf{Question.}\ In an equilateral triangle $ABC$ ($AB=1$), squares of side $c, 2c, 3c, 4c$ are placed side by side on $BC$ as a staircase, the first/last touching $AB$/$BC$. Find $c$ for $n=4$.

\smallskip
\textbf{Reference:}\ $\dfrac{6-\sqrt{3}}{55}$ \hfill \textbf{Model answer:}\ $\dfrac{6-\sqrt{3}}{55}$~\textcolor{trajsearch}{\faCheck}
\smallskip

\trturn{1}
\trquote{An equilateral triangle with a staircase of squares of sides $c, 2c, 3c, 4c$ on $BC$. Let me first look for this exact problem online.}
\trsearch{\texttt{equilateral triangle staircase squares side c 2c 3c nc}}

\trturn{2--9 (search fails)}
\trvisit{\textnormal{YouTube, Stack Exchange, cut-the-knot, Art of Problem Solving --- in turn}}
\trquote{None matches: the closest is a Stack Exchange post with only \emph{two} squares of sides $a, 2a$. The exact four-square staircase is not online --- I will have to derive it.}

\trturn{10--11 (adopt a method)}
\trvisit{\texttt{math.stackexchange.com/...\,two-squares}}
\trquote{That two-square solution forms a right triangle at each end with a $60^\circ$ angle and uses $\tan 60^\circ$. I can generalize the same construction to the four squares $c, 2c, 3c, 4c$.}

\trturn{14 (set up coordinates)}
\trquote{Set $B=(0,0)$, $C=(1,0)$, $A=(0.5, \tfrac{\sqrt3}{2})$. Then line $AB$ is $y=\sqrt3\,x$. With the square bases on $BC$, the staircase's outer corners must lie on $AB$; I write that contact condition together with the total-width constraint along $BC$.}

\trturn{15 (solve)}
\trquote{Using $\tan 60^\circ=\sqrt3$ for the two end triangles and summing the four square widths plus the end gaps to $1$ gives a single linear equation in $c$. Solving it yields $c = \dfrac{6-\sqrt3}{55}$.}
\transwer{$c = \dfrac{6-\sqrt{3}}{55}$ --- derived in closed form; no source online contained the answer.}
\end{trajbox}
\end{document}